\documentclass[10pt,journal,compsoc]{IEEEtran}
\usepackage{hyperref}

\hypersetup{colorlinks = true,linkcolor = blue,anchorcolor =red,citecolor = blue,filecolor = red,urlcolor = red,pdfauthor=author}

\usepackage{subfigure}
\usepackage{amsmath}

\usepackage[linesnumbered,ruled,vlined]{algorithm2e}

\usepackage{multirow}

\usepackage{tabularx}
\usepackage[table,xcdraw]{xcolor}
\usepackage{color}
\usepackage{soul}
\soulregister\cite7
\soulregister\ref7 

\usepackage{subfigure}

\usepackage{graphicx}

\usepackage{amsthm}
\usepackage{amssymb}

\usepackage{array} 

\ifCLASSOPTIONcompsoc
  \usepackage[nocompress]{cite}
\else
  \usepackage{cite}
\fi
\ifCLASSINFOpdf
\else
\fi
\begin{document}

%
\title{GAT-COBO: Cost-Sensitive Graph Neural Network for Telecom Fraud Detection}
%
%
%
%

\author{Xinxin~Hu,
        Haotian~Chen,
        Junjie~Zhang,
        Hongchang~Chen,
        Shuxin~Liu,
        Xing~Li,
        Yahui~Wang,
        and Xiangyang~Xue
\IEEEcompsocitemizethanks{\IEEEcompsocthanksitem Xinxin Hu, Junjie Zhang, Hongchang Chen, Shuxin Liu, Xing Li, and Yahui Wang are with National Digital Switching System Engineering and Technological Research Center, Zhengzhou, China, 450002.(E-mail:  hxx@alumni.hust.edu.cn, jj961004@stu.xjtu.edu.cn, ndscchc@139.com, liushuxin11@126.com, lixing\_ndsc@163.com, wangyahuindsc@163.com)\protect\\

\IEEEcompsocthanksitem Haotian Chen is with The Edward S. Rogers Sr. Department of Electrical \& Computer Engineering, University of Toronto, Toronto, Canada, M5S 3G4.(E-mail: hhaotian.chen@mail.utoronto.ca) \protect\\

\IEEEcompsocthanksitem Xiangyang Xue is with Institute of Big Data, Fudan University, Shanghai, China.(E-mail: xyxue@fudan.edu.cn)}

\thanks{This work is supported by Henan Province Major Science and Technology Project under Grant 221100210100, Central Plains Talent Foundation of China under Grant 212101510002 and the National Natural Science Foundation of China under Grant 61803384.}
\thanks{Corresponding author: Hongchang Chen, Shuxin Liu}
\thanks{Xinxin Hu and Haotian Chen contributed equally to this work and should be considered co-first authors.}

\thanks{Manuscript received XX XX, XXXX; revised XX XX, XXXX.}}

%
%

\markboth{IEEE Transactions on Big Data,~Vol.~14, No.~8, Oct~2022}%
{Shell \MakeLowercase{\textit{et al.}}: Bare Demo of IEEEtran.cls for Computer Society Journals}
%



\IEEEtitleabstractindextext{%
\begin{abstract}
Along with the rapid evolution of mobile communication technologies, such as 5G, there has been a drastically increase in telecom fraud, which significantly dissipates individual fortune and social wealth. In recent years, graph mining techniques are gradually becoming a mainstream solution for detecting telecom fraud. However, the graph imbalance problem, caused by the Pareto principle, brings severe challenges to graph data mining. This is a new and challenging problem, but little previous work has been noticed. In this paper, we propose a \underline{G}raph \underline{AT}tention network with \underline{CO}st-sensitive \underline{BO}osting (GAT-COBO) for the graph imbalance problem. First, we design a GAT-based base classifier to learn the embeddings of all nodes in the graph. Then, we feed the embeddings into a well-designed cost-sensitive learner for imbalanced learning. Next, we update the weights according to the misclassification cost to make the model focus more on the minority class. Finally, we sum the node embeddings obtained by multiple cost-sensitive learners to obtain a comprehensive node representation, which is used for the downstream anomaly detection task. Extensive experiments on two real-world telecom fraud detection datasets demonstrate that our proposed method is effective for the graph imbalance problem, outperforming the state-of-the-art GNNs and GNN-based fraud detectors. In addition, our model is also helpful for solving the widespread over-smoothing problem in GNNs. The GAT-COBO code and datasets are available at \textcolor{blue}{\url{https://github.com/xxhu94/GAT-COBO}}.
\end{abstract}

}

\maketitle

\IEEEdisplaynontitleabstractindextext

%
\IEEEpeerreviewmaketitle

\IEEEraisesectionheading{\section{Introduction}\label{sec:introduction}}

\begin{figure}[h]
  \includegraphics[width=\linewidth]{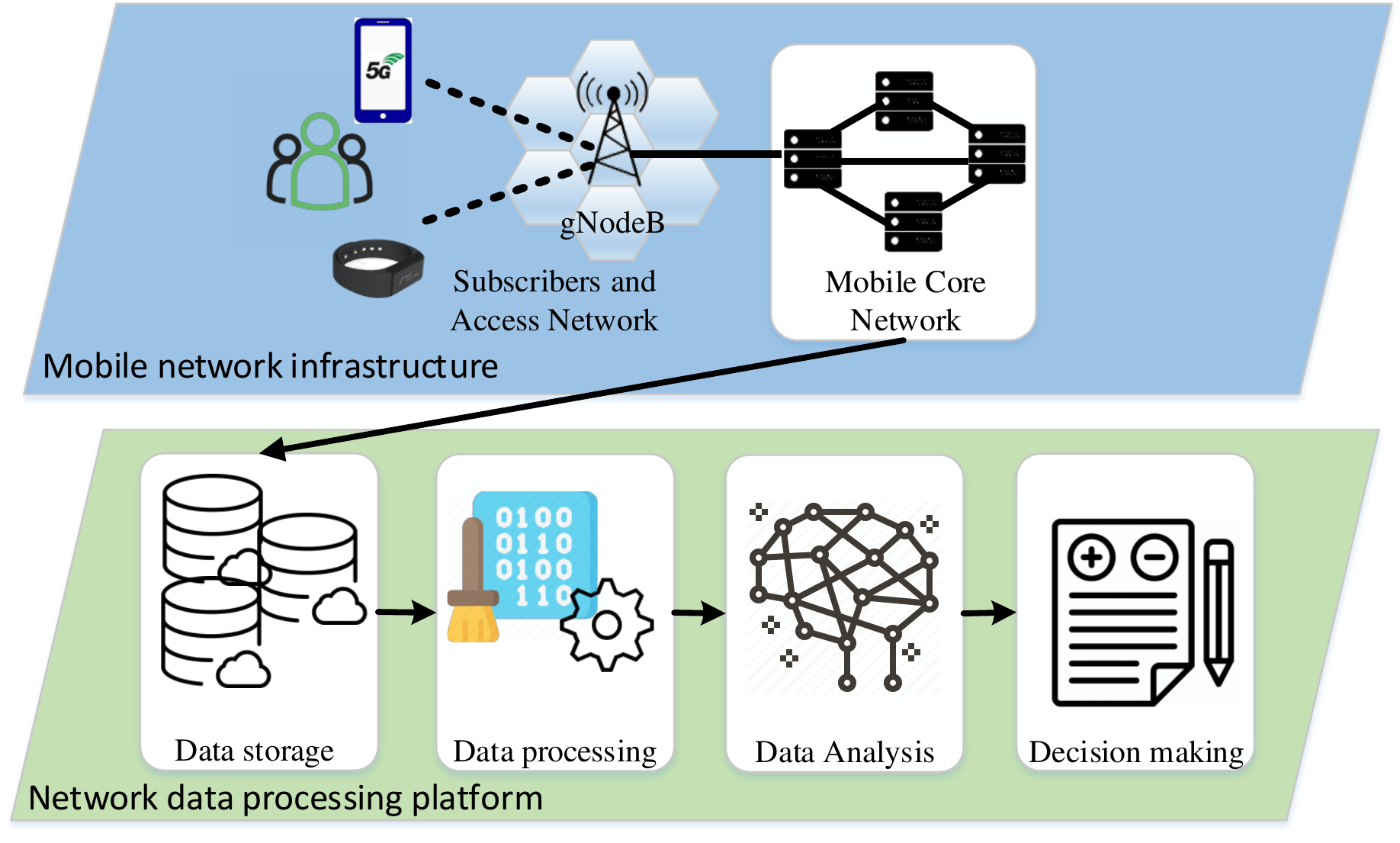}
  \caption{Illustration of mobile network big data platform and workflow.}
  \label{big data platform}
\end{figure}

\IEEEPARstart{T}{elecom} fraud has become increasingly rampant around the world over the past few years. In 2020, one-third of the U.S. population experienced telecom fraud, with losses of \$19.7 billion. In the same year, mainland China handled 230 million fraudulent phone calls and 1.3 billion fraudulent text messages \cite{caict2020fraud}. Not only is telecom fraud massive, but it is growing rapidly. Over the past six years, U.S. telecom fraud has grown at an average annual rate of 30\% \cite{truecaller2021spam}. Compared to 2019, mainland China also saw a 10\% increase in the number of telecom frauds disposed of in 2020. These scams can cause financial losses and even emotional trauma to a large number of victims \cite{tencent2022spam}. As the commercialization of 5G mobile networks progresses, the number of fraudsters and victims will inevitably increase further. How to unearth telecom fraudsters has become an increasingly important research topic.  Here, we take the operator's mobile network big data platform as an example to introduce telecom fraud and its detection pipeline (see Figure \ref{big data platform}). In an operator's mobile network infrastructure, subscribers and their devices generate a large number of network behavior records every day due to their communication behavior, many of which are generated by fraudulent users. With the help of network data processing platform, operators can mine user Call Detail Records (CDR) to detect fraudsters, thereby assisting mobile network operation decisions. Data analysis is the most crucial aspect of the whole process, and it is also full of challenges.

Subscribers' communication behaviors naturally constitute graphs, and the use of graph mining techniques for data analysis has become an important trend. In recent years, graph neural network (GNN) \cite{tseng2015fraudetector,dou2020enhancing, huang2022auc} has gradually become the mainstream technology for graph data mining. Current graph-based data mining tasks mainly model the relationships between nodes from the perspective of topology and attribute content \cite{zhang2020network}, making nodes of the same class more closely embedded in the embedding space and dissimilar nodes further away. A typical semi-supervised node classification task is performed as follows\cite{sun2009classification}: given a large graph with a small scale of node labels, a classifier is trained on those labeled nodes and used to classify other nodes during the testing process. These related works include graph convolutional networks (GCN) \cite{kipf2016semi} and many of its variants proposed in recent years \cite{zhang2019bayesian}, which effectively utilize features in the spectral domain by using simplified first-order approximations. GraphSage \cite{hamilton2017inductive} and Graph Attention Network(GAT) \cite{velivckovic2017graph} utilize features in the spatial domain to better adapt to different graph topologies. GNNs have achieved remarkable performance in many application domains, such as text classification \cite{yao2019graph}, image recognition \cite{chen2019multi} , and recommender systems \cite{wang2019knowledge}. GNN-based graph data anomaly detection has also made great progress \cite{ma2021comprehensive}. 

However, a significant challenge in graph anomaly detection is the imbalance between normal and abnormal classes. For example, only a tiny fraction of the huge amount of CDR data stored by telecom operators are fraudulent users. The same is true of fraudulent users in social networks and financial transaction networks. Existing graph anomaly detection methods usually assume that the class distribution in the input graph data is almost or completely balanced, or deliberately provide each class with a balanced labeled sample at training time. This ensures that the representation across multiple classes is balanced, thus completely avoiding the class imbalance problem. However, this artificial balance interference is obviously inconsistent with the distribution of real-world graph data. Because different parts in real-world graph-based systems often evolve asymmetrically and unrestrictedly, which makes the data naturally exhibit highly skewed class distributions. If the imbalance problem is not considered when designing the GNN model, the majority class may dominate the loss function. This makes the trained GNN overclassify these majority classes and fails to accurately predict samples from the minority class, which are the real focus of our attention. When generalizing models to graphs with imbalanced class distributions, existing GNN methods tend to overfit the majority class, resulting in suboptimal embedding results for the minority class.

In the field of machine learning, in order to solve the class imbalance problem, researchers mainly adopt data-level methods, algorithm-level methods, and hybrid methods \cite{johnson2019survey}. Data-level methods seek to operate on the data to make the class distribution more balanced, such as over-sampling or under-sampling \cite{chawla2002smote,drummond2003c4}. Algorithm-level methods typically introduce different misclassification penalties or prior probabilities for different classes \cite{dong2018imbalanced,ling2008cost,zhou2005training,khan2017cost}. Hybrid methods \cite{sun2007cost,ando2017deep,chawla2003smoteboost} attempt to combine the above two. However, applying these methods directly to graph data mining may not yield optimal results. Because the vast majority of existing work on the imbalance problem is devoted to $i.i.d$ data. The relational characteristics of graph data are obviously in conflict with the $i.i.d$ assumption. In addition, GNN also has an over-smoothing problem. When the number of GNN layers is too large, the features learned by all nodes tend to be consistent, resulting in a sharp decline in the classification performance of the model, which puts forward more stringent conditions for the application of unbalanced methods to graph data mining.

 In order to solve the above problems, we combine GAT and ensemble learning to design a cost-sensitive GNN. Specifically, we first treat each GAT as a weak classifier and use it to learn the embedding representation for each node. Then the embedding is input to the cost-sensitive learner, which will calculate the classification bias. According to the obtained deviation, the misclassification cost and the updated node sampling weight are calculated, so the misclassified samples are given greater weight in the next weak classifier. Then, we use the updated weights to constrain the loss function of the next weak classifier, thus to retrain to obtain new embedding representations. By concatenating such multiple weak classifiers and summing the embedding, the final cost-sensitive embedding representation is obtained.

Based on the above ideas, we propose a new model that performs anomaly detection by boosting GNN with cost-sensitive learning. Our contributions are as follows:
\begin{itemize}
\item
We reveal the graph imbalance problem in telecom fraud and design a novel semi-supervised GNN framework for its detection.
\item
Combining Boosting and GNN, we try to embed GAT as a base classifier in the ensemble learning framework, which not only improves GNN performance but also overcomes the over-smoothing effect.
\item
We design a cost-sensitive learning scheme for GNN to solve the graph imbalance problem and provide a theoretical proof.
\item
Extensive experiments are conducted on two real-world telecom fraud datasets to demonstrate the effectiveness of the proposed method.
\end{itemize}


The rest of the paper is organized as follows. Sec.\ref{related work} reviews the related works. Sec.\ref{problem definition} introduces definitions and the problem statement. In Sec.\ref{method}, we give the details of GAT-COBO. In Sec.\ref{experiment}, we conduct experiments to evaluate the effectiveness of GAT-COBO. In Sec.\ref{conclusion}, we conclude the paper.

\section{Related work}\label{related work}

\subsection{GNN-based fraud detection}
In recent years, GNN, with its powerful graph data representation ability, has been widely used in fraud detection tasks. According to the application fields, these works can be divided into three main categories, namely, GNN-based telecom fraud detection  \cite{tseng2015fraudetector,jiang2022telecom,liu2019agrm,ji2020multi,zheng2018generative,irarrazaval2021telecom,ravi2022wangiri},  GNN-based social network fraud detection \cite{dou2020enhancing,li2019spam,liu2020alleviating,wang2019fdgars,liu2022nowhere}, and GNN-based financial fraud detection \cite{li2022ttagn,ao2021temporal,liang2021credit,ji2022prohibited,liu2018heterogeneous,wang2019semi,zhao2020error,zhong2020financial}.  

(1) Telecom fraud detection. Liu et al. \cite{liu2019agrm} and Ji et al. \cite{ji2020multi} used attention mechanism based graph neural network and Multi-Range Gated Graph Neural Network (MRG-GNN) for telecom fraud detection, respectively. Based on the constructed directed bipartite graph, Tseng et al. \cite{tseng2015fraudetector} learned the trust values of remote phone numbers by a weighted HITS algorithm. Zheng et al. \cite{zheng2018generative} proposed a generative adversarial network (GAN) based model to calculate the probability that a bank transfer is a telecom fraud. Recently, Jiang et al. \cite{jiang2022telecom} detected telecom fraud by integrating the Hawkes process into an LSTM for historical impact learning. Based on the subscriber's CDR and payment record information, Chadysas et al. \cite{chadyvsas2022outlier} used univariate outlier detection methods to identify fraudulent customers in mobile virtual network operators (MVNOs). Krasic et al. \cite{krasic2022telecom} combined machine learning and SMOTE oversampling to solve the highly unbalanced data distribution problem in telecom fraud detection. Yang et al. \cite{yang2019mining} disclosed the different behavioral characteristics of fraudsters and non-fraudsters in mobile networks, and designed a semi-supervised detection model based on factor graphs.

(2) Social network fraud detection. Dou et al. proposed CARE-GNN \cite{dou2020enhancing}, which augments the aggregation process of GNN with reinforcement learning to prevent fraudsters disguise for opinion fraud detection. With heterogeneous and homogeneous graph-based neural networks, Li et al. proposed GAS \cite{li2019spam} to capture local context and global context information of comments. GraphConsis \cite{liu2020alleviating} studied the inconsistency of context, features and relationships in graph-based fraud detection. FdGars \cite{wang2019fdgars} employed graph convolutional networks for fraud detection in online application review systems. Liu et al. \cite{liu2022nowhere} proposed a novel rumor detection framework based on structure-aware retweeting graph neural network. 

(3) Financial fraud detection.
Li et al. \cite{li2022ttagn} proposed Temporal Transaction Aggregation Graph Network (TTAGN) for Ethereum phishing fraud detection. Ji et al. \cite{ji2022prohibited} introduced structural learning into large-scale risk graphs to solve the problem of prohibited item detection in e-commerce. Ao et al. \cite{ao2021temporal} proposed to detect fraudulent accounts on transaction graphs constructed from Ethereum block transactions. GEM \cite{liu2018heterogeneous} adaptively detected malicious accounts from heterogeneous account device graphs. Wang et al. proposed Semi-GNN \cite{wang2019semi}, a hierarchical attention GNN,  for financial fraud detection. Unlike making changes in the GNN model, Zhao et al. \cite{zhao2020error} designed a graph anomaly loss function for training the anomaly node representation of GNNs. Besides, Zhong et al. proposed MAHINDER \cite{zhong2020financial}, which explores metapath on heterogeneous information networks of multi-view attributes for credit card fraud transaction detection. Saia et al. \cite{saia2019evaluating} argued the advantages of using proactive fraud detection strategies over traditional retrospective strategies. 

Furthermore, a comprehensive review of graph-based fraud detection techniques was provided by Pourhabibi et al \cite{pourhabibi2020fraud}. Ma et al. \cite{ma2021comprehensive} conducted a systematic and comprehensive review of contemporary deep learning techniques for graph anomaly detection.

\subsection{Class imbalance learning} 
Class imbalance learning is an important research direction in the field of data mining and machine learning \cite{sun2009classification,johnson2019survey,buda2018systematic,he2009learning}. In practice, classes with a large number of instances are often called majority classes, and classes with fewer instances are often called minority classes. The methods to solve the class imbalance problem can be roughly divided into three categories, namely data level, algorithm level and hybrid level.

(1) Data-level methods try to rebalance the previous class distribution through a preprocessing step. There are two data-level approaches. One is to oversample the minority class, such as SMOTE \cite{chawla2002smote}, which solves this problem by generating new samples, and performing interpolation between the minority class samples and their nearest neighbors. But oversampling methods can lead to overfitting. Another approach is to undersample the majority class \cite{drummond2003c4}, but this may discard valuable information. 

(2) Algorithm-level approaches try to modify existing algorithms to emphasize minority classes, such as designing new loss functions, or specifying cost matrices for models (also known as cost-sensitive learning \cite{dong2018imbalanced}). Cost-sensitive learning \cite{ling2008cost,zhou2005training} usually builds a cost matrix to assign different misclassification penalties to different classes. Khan et al. \cite{khan2017cost} proposed a method that can automatically optimize the cost matrix by backpropagation. Lin et al. \cite{lin2017focal} balanced the model by designing a loss function with a threshold (Focal loss) to increase the loss weight of the minority class samples and reduce the weight of the majority class samples. 

(3) The last category is hybrid methods, which combine the data-level and algorithm-level approaches.  Ando et al. \cite{ando2017deep} proposed the first deep feature oversampling method. Chawla et al.\cite{chawla2003smoteboost} combined boosting with the SMOTE method. The works \cite{sun2009classification,johnson2019survey,buda2018systematic,he2009learning} also provided a review of solutions to the class imbalance problem.

In the field of graph learning, graph imbalance is a novel problem that only a few works have considered. For example, an early work \cite{bertoni2011cosnet} proposed a Hopfield-based cost-sensitive neural network algorithm (COSNet). PC-GNN \cite{liu2021pick} performed imbalanced graph learning by oversampling and undersampling different category nodes. Graphsmote \cite{zhao2021graphsmote} combined SMOTE sampling algorithm and GNN to solve the graph imbalance problem. DR-GCN \cite{shi2020multi} employed conditional adversarial training and distribution alignment to learn robust node representations of majority and minority classes. 

However, these efforts are mainly to solve the imbalance problem in graph by sampling methods. As mentioned above, sampling methods have many limitations. There is no work yet to combine cost-sensitive learning with modern GNNs. In this work, we consider introducing cost-sensitive learning into graph neural networks in an ingenious way to efficiently address the graph imbalance problem.

\begin{table}
  \caption{Glossary of Notations}
  \label{tab:commands}
  \begin{tabular}{cl}
    \hline
    Symbol &Definition \\
    \hline
     $\mathbf{A}$ & Adjacency matrix \\
     $\mathbf{X}^{(l)}$ & Input feature matrix of the $l$-th layer, $\mathbf{X}^{(0)}=\mathbf{X}$ \\
     $\alpha_{ij}$ & Attention coefficient between nodes $i$ and $j$ \\
     $\mathbf{\Omega}^{(l)}$ & The attention matrix of layer $l$ \\
     $\mathbf{W}$ & Weight matrix of neural network \\
     $\mathbf{h_v}$ &The embedding of node $v$ in the  GAT classifier \\
     $\mathbf{z}_v$ & Final embedding of node $v$ in the GAT classifier \\
     $p_{k}^{(l)}(v)$ & Probability of $v$ classified as class $k$ in $l$-th GAT classifier\\
     $h_{k}^{(l)}(v)$ & Class $k$ probability of node $v$ in $l$-th cost-sensitive learner  \\
     $w_{v}^{l}$ & The sample weight of node $v$ at layer $l$ \\
     $\mathbf{C}$ & Cost sensitive matrix \\
     $C_v^{(l)}$ & The misclassification cost of node $v$ at layer $l$ \\
     $C_{ij}$ &Cost of classifying a node in class $i$ as the class $j$\\
     $D^{(l)}(x_i)$ & The distribution of node $x_i$ in the $l$-th layer\\
    \hline
  \end{tabular}
  \label{notation}
\end{table}

\section{Problem Definition}\label{problem definition}

In this section, we introduce the definition of Graph, Graph imbalance problem, cost-sensitive learning, and GNN-based telecom fraud detection. Table \ref{notation} summarizes all the basic symbols.

\noindent\textbf{Definition 3.1. Graph.} In general, a graph can be defined as $\mathcal{G}=(\mathcal{V},\mathcal{X},\mathcal{A},\mathcal{E},\mathcal{Y})$ , where $\mathcal{V}=\{ v_1, v_2, v_3,...v_N \}$ is a set of nodes. $\mathcal{X}=\{\mathbf{x}_1,\mathbf{x}_2,...\mathbf{x}_N \}$ is the set of node features, where $\mathbf{x}_i  \in  \mathbb{R} ^d$ is the feature vector of node $v_i$. Stacking these vectors into a matrix constitutes the feature matrix $\mathbf{X} \in \mathbb{R}^{N \times d}$ of the graph $\mathcal{G}$. $\mathbf{A} \in \mathbb{R}^{N \times N}$ represents the adjacency matrix of $\mathcal{G}$, where $a_{i,j}=1$ means that there is an edge between node $v_i$ and node $v_j$, if not, $a_{i,j}=0$. $\mathcal{E}$ represents the set of edges, that is, $ \mathcal{E}=\{ e_1, e_2, e_3,...e_M \}$. $e_j=(v_{s_j},v_{r_j}) \in \mathcal{E}$ is an edge between node $v_{s_j}$ and $v_{r_j}$, where $v_{s_j}, v_{r_j} \in \mathcal{V}$. $\mathcal{Y}=\{y_1,y_2,...y_N \}$ is the set of labels corresponding to all nodes in the set $\mathcal{V}$. For the convenience of representation, we encode the label $y_i$ as one-hot vector $\mathbf{y}_i$.

\noindent\textbf{Definition 3.2. Graph imbalance problem.} Given the labels $\mathcal{Y}$ of a set of nodes in a graph $\mathcal{G}=(\mathcal{V},\mathcal{X},\mathcal{A},\mathcal{E},\mathcal{Y})$, there are $K$ classes in $\mathcal{Y}$, namely $C=\{C_1, ...,C_K \}$.  $|C_{i}|$ is the size of the $i$-th class, that is, the number of samples belonging to class $i$. We use
\begin{equation}
IR=\frac{min_i(|C_i|)}{max_i(|C_i|)}
\end{equation}
to measure the class imbalance ratio. Therefore, $IR$ lies in the range $[0, 1]$ . The smaller the $IR$, the more severe the imbalance \cite{he2009learning}. In particular, in a binary classification problem, $C_1$ and $C_2$ denote two classes in $C$ , where $C_1$ is the minority class , $C_2$ is the majority class. Then the imbalance ratio of $C_1$ and $C_2$ is defined as $IR=|C_1|/|C_2|$.

\noindent\textbf{Definition 3.3. Cost-sensitive learning.} Cost-sensitive learning mainly considers how to train a classifier when different classification errors lead to different penalties. In scenarios such as fraud detection, medical diagnosis, network security, etc., misclassifying the minority class often leads to large losses. Different from traditional classification methods to minimize the misclassification rate, cost-sensitive learning mainly introduces misclassification costs into classification decisions to reduce the overall cost of mis-classification.

Without loss of generality, given a set of labels $C$, there are $K$ classes in total. Cost-sensitive learning measures the misclassification loss of the machine learning algorithm by defining the misclassification cost. Specifically, the misclassification cost matrix can be defined as the following form:
\begin{align}
\mathbf{C} & = \left[\begin{array}{cccc}
C_{10} & C_{11} & \cdots & C_{1 K} \\
C_{20} & C_{21} & \cdots & C_{2 K} \\
\vdots & \vdots & \ddots & \vdots \\
C_{K 0} & C_{K 1} & \cdots & C_{K K}
\end{array}\right]
\label{misclassification}
\end{align}
where $C_{ij}$ represents the misclassification cost of classifying the sample in class $i$  as class $j$. And the larger the value, the greater the loss caused by the misclassification. $C_{ij} \subset [0, +\infty)$ is an associated cost item. In cost-sensitive learning, the cost matrix $\mathbf{C}$ can be set in advance by domain experts based on experience, or obtained through parameter learning. In cost-sensitive learning, the cost matrix $\mathbf{C}$ varies with specific application problems, and it can generally be determined by parameter learning methods or by domain experts based on experience. At the same time, $\mathbf{C}$ should generally satisfy the constraint: the values of one row in the cost matrix cannot all be greater than the values of another row. For example, for row $m$ and $n$ $(1\leq m,n \leq K)$ of $\mathbf{C}$, if $C_{mj}>C_{nj}$ holds in all columns $j$, then the prediction result of the test sample is always class $n$ according to minimum expected misclassification cost criteria.

\noindent\textbf{Definition 3.4. Graph-based fraud detection.} With subscribers in the operator's network as nodes, subscribers behavior as node features, and communication between subscribers as edges to build graph, we can then use GNN for fraudulent subscriber detection. Specifically, for a given subscriber behavior graph $\mathcal{G}=(\mathcal{V},\mathcal{X},\mathcal{A},\mathcal{E},\mathcal{Y})$, the inter-layer transfer formula in the GNN-based telecom fraud detection model can be formally described as:
\begin{equation}
 \mathbf{h}_{v}^{(l)}=\sigma\left(\mathbf{h}_{v}^{(l-1)} \oplus \operatorname{Agg}\left(\left\{\mathbf{h}_{v^{\prime}}^{(l-1)}:\left(v, v^{\prime}\right) \in \mathcal{E} \right\}\right)\right.
\label{inter-layer transfer formula}
\end{equation}
where $\mathbf{h}_v^{(l)}$ represents the embedding of node $v$ at layer $l$, $\mathbf{h}_v^{(0)}=\mathbf{x}_v$, and $v^{\prime}$ is a neighbor of node $v$. $Agg()$ is a message aggregation function for neighbor feature aggregation, such as mean aggregation, pooling aggregation, attention aggregation, etc. $\oplus$ represents feature concatenation or summation operation.

\section{The Proposed Method}\label{method}

\begin{figure*}[h]
  \includegraphics[width=\linewidth]{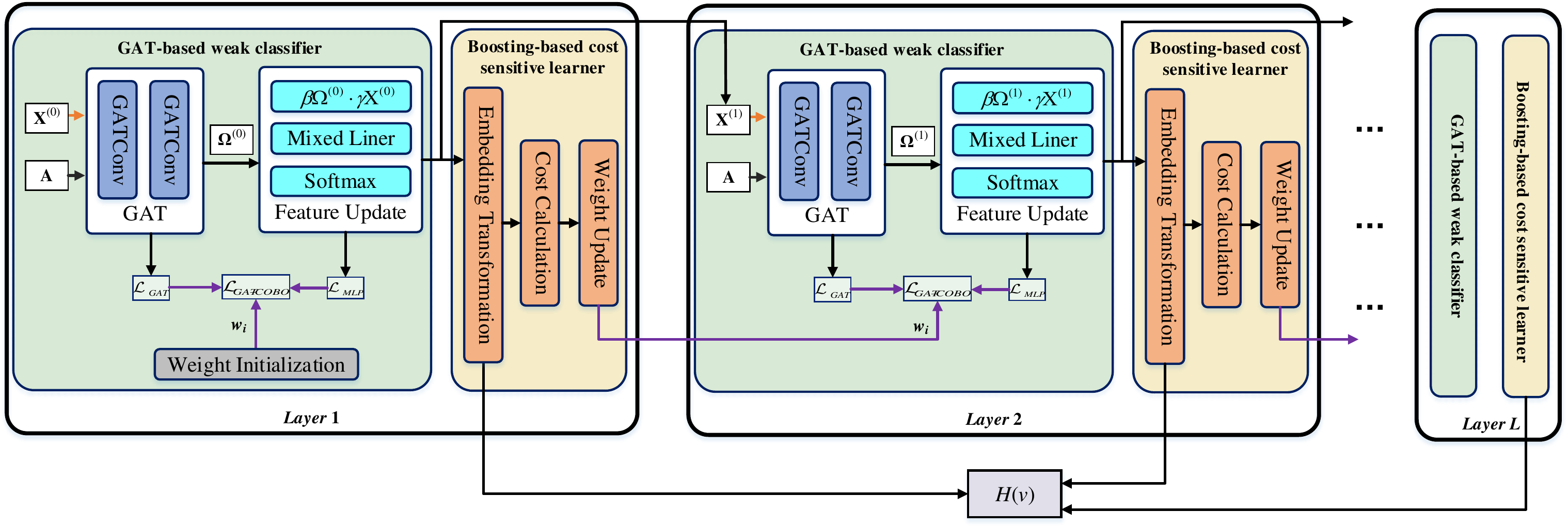}
  \caption{The overview of our proposed model GAT-COBO.}
  \label{deep fraudsters}
\end{figure*}

\subsection{Overview}\label{Overview}
Our proposed model consists of two main parts, one is a GAT-based weak classifier and the other is a boosting-based cost sensitive learner. We illustrate the pipeline of the proposed method in Figure \ref{deep fraudsters}. To begin with, the node features and adjacency matrix of the graph are fed into a GAT-based weak classifier which can be trained to obtain the node embeddings by signals from the labels (Section \ref{GNN-based weak classifier}). Then, we feed the embeddings into the cost-sensitive learner to compute the misclassification cost of this weak classifier and update the sampling weights of the corresponding nodes according to it. After that, the updated sampling weights are used to guide the training of the next GAT-based weak classifier. Finally, the embeddings of multiple concatenated weak classifiers are summed to obtain the final embedding of every node which can be used for node classification in downstream tasks (Section \ref{Boosting-based cost sensitive learner}). The entire optimization steps and algorithm are presented in Section \ref{Propose GAT-COBO}.

\subsection{GAT-based weak classifier}\label{GNN-based weak classifier}
To improve the performance of GNNs, previous studies have tended to explore in terms of information transformation and aggregation approaches of GNNs \cite{you2020design}. But these models are difficult to work on graph imbalanced data. Here, we try to change the perspective and use the classical GAT as a base weak classifier for the proposed method. The reason for choosing it is that GAT, as a classic GNN model, has strong expressive ability and can learn different attention weights for different neighbors, which is critical for the imbalance problem. For example, the GNN that generally uses the $mean()$ function as the aggregator can be regarded as a simplification of GAT (equivalent to setting the attention coefficient of all connected nodes to 1). There are two modules in each GAT-based weak classifier, namely the graph attention module and the feature update module. The design of these two modules is elaborated below.

\subsubsection{Graph attention module}
Given an undirected graph $\mathcal{G}=(\mathcal{V},\mathcal{X},\mathcal{A},\mathcal{E},\mathcal{Y})$, according to vanilla GAT in \cite{velivckovic2017graph}, the attention coefficient between adjacent nodes can be expressed as follows:
\begin{equation}
\alpha_{i j}=\frac{\exp \left(\text { LeakyReLU }\left(\mathbf{a}^{T}\left[\mathbf{W} \mathbf{h}_{i} \| \mathbf{W h}_{j}\right]\right)\right)}{\sum_{k \in \mathcal{N}_{i}} \exp \left(\operatorname{LeakyReLU}\left(\mathbf{a}^{T}\left[\mathbf{W} \mathbf{h}_{i} \| \mathbf{W h}_{k}\right]\right)\right)}
\label{attention coefficients1}
\end{equation}
where $\mathbf{a}$ represents the attention function to be learned and is implemented by a single-layer MLP.  $\mathbf{W} \in \mathbb{R}^{hid \times d}$ represents the linear transformation of the node embedding. $hid$ is the number of neurons, which is set manually as a model hyper-parameter, and $d$ is the dimension of $\mathbf{h}$. $||$ represents the feature concatenation, and $LeakyRelu()$ is the activation function.

Therefore, the new embedding $\mathbf{h}_i^{\prime}$ of node $i$  is calculated according to the attention coefficient and the old embedding $\mathbf{h}_i$ as follows:
\begin{equation}
\mathbf{h}_{i}^{\prime}=\sigma\left(\sum_{j \in \mathcal{N}_{i}} \alpha_{i j} \mathbf{W h}_{j}\right)
\label{innerGAT embedding1}
\end{equation}
where $\mathcal{N}_{i}$ is the set of neighbors of node $i$. $\sigma$ is activation function.
 
In order to obtain a richer representation, a multi-head attention mechanism can be used as in \cite{velivckovic2017graph}:
\begin{equation}
\mathbf{h}_{i}^{\prime}=\|_{q=1}^{Q} \sigma\left(\sum_{j \in \mathcal{N}_{i}} \alpha_{i j}^{q} \mathbf{W}^{q} \mathbf{h}_{j}\right)
\label{innerGAT embedding2}
\end{equation}
where $Q$ represents the number of heads, and $\alpha_{ ij }^{q}$ represents the attention coefficient of the $q$-th head.

To train the GAT with the signals from the labels, we adopt the minimization of the cross-entropy loss function as the optimization objective:
\begin{equation}
\mathcal{L}_{GAT}=-\sum_{v \in \mathcal{V}}\left(\mathbf{y}_{v} \log \mathbf{z}_{v}^{(l)}\right)
\label{inter loss}
\end{equation}
where $\mathbf{z}_v ^{(l)} = (\mathbf{h}_{v}^{\prime})_{final} $ is the final embedding of node $v$ for the $l$-th weak classifier. 

By supervising the training process with the above loss function, the optimal attention function is learned. So, the attention coefficients of all nodes can be obtained to form the attention matrix as follows:
\begin{equation}
\mathbf{\Omega}=\left[\begin{array}{cccc}
\alpha_{11} & \alpha_{12} & \cdots & \alpha_{1 N} \\
\alpha_{21} & \alpha_{22}& \cdots & \alpha_{2N}\\
\vdots & \cdots& \cdots & \alpha_{kk} \\
\alpha_{N 1} & \cdots & \cdots & \alpha_{N N}
\end{array}\right]
\label{Omega}
\end{equation}
where ${\alpha}_{ij}$ is calculated through Eq. (\ref{attention coefficients1}) if $v_i$ and $v_j$ are neighbors, otherwise ${\alpha}_{ij}=0$.

\subsubsection{Feature update module}
The Boosting algorithm requires concatenation between weak classifiers \cite{Freund1997A}. However, it is a novel challenge to implement multiple GAT-based classifiers in concatenation. An intuitive idea is to use the final embedding of nodes from the previous GAT weak classifier as the initial feature input to the next weak classifier. Unfortunately, this design is approximately equivalent to increasing the number of GAT layers, which not only fails to improve the model performance, but also leads to over-smoothing effect of the GNN model. As pointed out in \cite{oono2019graph}, after stacking multiple layers in the GNN model, the nodes aggregate too many hops of neighborhood information, leading to a convergence of features of all nodes. Thus, the unique feature information of each node is lost and only the structural information remains, thus leading to an over-smoothing effect. To solve this problem, we include the original feature information $\mathbf{X}$ in the input of each base classifier along with the attention matrix $\mathbf{\Omega}$ learned in the previous base classifier, which avoids the loss of node feature information due to too deep GNN layers and also utilizes the attention matrix learned in the previous base classifier. Based on the above considerations, we design a feature update module for each GAT weak classifier.This module can feed the attention information learned in the $(l-1)$th base classifier into the $l$-th one. In this method, the input feature of the $l$-th weak classifier becomes:

\begin{equation}
\mathbf{X}^{(l)}=\left(\beta \mathbf{\Omega}^{(l-1)}\right) \cdot\left(\gamma \mathbf{X}^{(l-1)}\right)
\label{input feature}
\end{equation}
where $\beta$ is a parameter used to constrain the attention weight, $\gamma$ is a parameter used to constrain the input feature. It is beneficial to update features in this way, because effective information transfer between classifiers can be easily achieved by passing the attention matrix of the previous weak classifier to the next one. The next weak classifier can be fine-tuned based on the previous weak classifier, thus ensuring that the results of the last correct classification are retained, while focusing more on the nodes that were misclassified in the last iteration.

To facilitate Adaboost ensemble learning, we input the node embeddings learned by each weak classifier into a mixed linear layer. A softmax operation is then performed to obtain the probability of each node being assigned to each class. In this method, the probability that the node $v$ of the $l$-th weak classifier belongs to the $k$-th class can be formally described as:
\begin{align}
p_{k}^{(l)}(v) &=Softmax\left[Mixliner\left(\mathbf{x}_v^{(l)}\right)\right]  \nonumber\\
&=\frac{\exp \left(\left[\operatorname{Relu}\left(\mathbf{W}^{(l)} \mathbf{x}_v^{(l)} \right)\right]_{k}\right)}{\sum_{k^{\prime}=1}^{K} \exp \left(\left[\operatorname{Relu}\left(\mathbf{W}^{(l)} \mathbf{x}_v^{(l)} \right)\right]_{k^{\prime}}\right)}
\label{pkv}
\end{align}
where $\mathbf{x}_v^{(l)}$ is the updated feature of node $v$ from updated feature matrix $\mathbf{X}^{(l)}$ defined in Eq. (\ref{input feature}). $Relu()$ is activation function. $\mathbf{W}$ represents the weight of the neural network to be learned, which is an independent weight parameter and does not share with the weights in Eq. \ref{attention coefficients1}, \ref{innerGAT embedding1} and \ref{innerGAT embedding2}.

The probability $p_{k}^{(l)}(v)$ of each node $v$ on $K$ different classes constitutes a probability vector $\mathbf{p}^{(l)}(v) \in \mathbb{R}^{K}$. We use the cross-entropy loss function to train the above process.
\begin{equation}
\mathcal{L}_{MLP}=-\sum_{v \in V}\left(\mathbf{y}_{v} \log \mathbf{p}^{(l)}(v)\right)
\label{cross-entropy loss}
\end{equation}

\subsection{Boosting-based cost sensitive learner}\label{Boosting-based cost sensitive learner}
Boosting is a cluster of simple, effective and well-explained ensemble learning methods, the best known of which is Adaboost \cite{Freund1997A}. Researchers have made many improvements to Adaboost since it was proposed. One of them is SAMME.R algorithm \cite{hastie2009multi}, a variant of Adaboost with fast convergence and high classification accuracy. In addition, to make Adaboost adaptable to the imbalance problem, Zhang et al. \cite{1999AdaCost} proposed a cost-sensitive Adaboost. But how to combine cost-sensitive learning with SAMME.R algorithm, especially GNN-based weak classifier, is a new problem that has hardly been explored before. In this paper, we design a cost-sensitive algorithm based on SAMME.R to solve the graph imbalance problem in GNNs, which we call Boosting-based cost-sensitive learner. Its structure is shown in Figure \ref{deep fraudsters}. Now, we elaborate on the design details.

\subsubsection{Boosting GNN with cost}

The SAMME.R algorithm requires that the output probability vector of each weak classifier needs to satisfy certain constraints in order for the final integrated result to be optimal. So we transform the node class probability vectors obtained in Equation (\ref{pkv}) according to the algorithm in \cite{hastie2009multi}. Given the output $p_k^{(l)}(v)$ of the $l$-th GAT-based weak classifier, we can calculate the node class probability of the $l$-th cost-sensitive learner $h^{(l)}(v)$:

\vspace{-0.5cm}
\begin{equation}
    \begin{aligned}
h_{k}^{(l)}(v) = & (K-1)\bigg(\log p_{k}^{(l)}(v) \bigg. \\
                 & \bigg. -\frac{1}{K} \sum_{k^{\prime}=1}^{K} \log p_{k^{\prime}}^{(l)}(v)\bigg),  k=1, \ldots, K
    \end{aligned}
\label{weak classifier}
\end{equation}

Based on the result, we can get the predicted label of node $v$ as $\arg \max _{k}(h_{k}^{(l)}(v))$ in the $l$-th weak classifier. Subsequently, we can calculate the misclassification cost of node $v$ as:
\begin{equation}
C_v^{(l)}=\mathbf{C}[y_v,\arg \max _{k}(h_{k}^{(l)}(v))]
\label{Cvpred}
\end{equation}
where $\mathbf{C}$ is the cost matrix defined in Sec.\ref{problem definition}.

The core idea of the Boosting algorithm is to combine multiple weak classifiers into one strong classifier. One of the most important steps to achieve this design is the weights updating between different weak classifiers. In order to make the classifier continuously optimize the classification results during training, the boosting algorithm assigns weight $w_v$ to each node $v$. If the classification result of node $v$ is wrong in the previous base classifier $h_{k}^{(l)}$, its weight $w_v$ is increased in the next base classifier $h_{k}^{(l+1)}$, so that node $v$ is valued more in $h_{k}^{(l+1)}$ and its probability of correct classification is increased. Otherwise, $w_v$ is adjusted downward. $w$ has an initial value of $1/N$. Also considering the class imbalance problem in the graph, we introduce cost-sensitive factors here. According to the classification result of $v$ in the weak classifier $l$, the higher the misclassification cost, the higher the weight of the node in the next weak classifier. Thus, the node weight update can be described formally as follows:

\vspace{-0.5cm}
\begin{equation}
w_{v}^{(l+1)}=w_{v}^{(l)} \cdot \exp \left(-\frac{K-1}{K} \cdot C_{v}^{(l)} \cdot \mathbf{y}_{v} \cdot \log \mathbf{p}^{(l)}(v)\right), v \in \mathcal{V}
\label{node weight update}
\end{equation}
where $w_{v}^{(l)}$ represents the weight of node $v$ at classifier $l$, $C_{v}^{(l)}$ represents the misclassifying cost of $v$ calculated through Eq.(\ref{Cvpred}).

\subsubsection{Cost matrix calculation}
While we give the form of the cost matrix in Section \ref{problem definition}, we do not specify how its value is calculated. However, the determination of the cost matrix is crucial for cost-sensitive learning. In practical problems, the cost matrix can be manually specified based on the actual loss caused by the sample being misclassified. However, the calculation of the cost matrix is still very challenging in many cases. In order to facilitate the calculation intuitively and ensure the reasonableness of the cost matrix, we elaborate on three general cost matrix calculation methods from the perspective of the sample imbalance rate $IR$, which are \textit{Uniform}, \textit{Inverse} and \textit{Log1p}, respectively.

\noindent\textbf{Uniform:} Set the misclassification cost of each class to be the same as 1. Such a cost matrix does not treat different classes differently. As a variant of the proposed method, it can be used when the classes are balanced, and it can also be used to compare the difference in the performance of the proposed model when classes are unbalanced.

\noindent\textbf{Inverse:} Set the misclassification cost to be the inverse of the sample class ratio. The cost $C_{ij}$ that the sample of the actual class $i$ is predicted to be $j$ can be expressed as:
\begin{equation}
C_{i j}=\frac{|C_j|}{|C_i|}
\label{actual class}
\end{equation}
where $C_j$ represents the predicted class and $C_i$ the actual class. $|C_i|$ represents the number of samples of class $i$. In this way, the cost of predicting the minority class as the majority class will be greater than 1, and the more unbalanced the sample, the higher the cost.

However, the \textit{Inverse} variant has an obvious drawback, that is, when the samples are highly unbalanced, $C_{ij}\rightarrow +\infty$, and the approximation speed is proportional to the sample unbalance ratio. Under such circumstances, the model will predict all majority classes as minority classes, resulting in low scores on metrics like Precision and F1. To reduce the approximation speed of $C_{ij}$ , we propose another scheme.

\noindent\textbf{Log1p:} Perform the $log1p()$ operation on the sample class ratio $C_{ij}$ in the \textit{Inverse} variant, which can make the biased cost values in the cost matrix conform to the unbiased normal distribution. And the formal expression is as follows:
\begin{equation}
C_{i j}=\log 1 p\left(\frac{|C_j|}{|C_i|}\right)=\log \left(\frac{|C_j|}{|C_i|}+1\right)
\label{log1p}
\end{equation}

After this operation, the cost matrix $\mathbf{C}$ will satisfy the constraints in Section \ref{problem definition}. Even if the sample class ratio is relatively unbalanced, extreme cost values are less likely to occur.

By choosing any one of the three cost value calculation methods above, we can update the node weights in the weak classifier with the help of Eq. (\ref{node weight update}). Combining the weak classifiers $h^{(l)}$ of each layer, the ensemble classification result can be calculated as follows:
\begin{equation}
H(v)=\arg \max _{k} \sum_{l=1}^{L} h_{k}^{(l)}(v)
\label{ensemble classification}
\end{equation}
where $h_{k}^{(l)}(v)$ is obtained through Eq. (\ref{weak classifier}).

\subsubsection{Theoretical proof}
For the Boosting algorithms, the choice of weight update parameters is crucial for converting a weak learning algorithm to a strong one. When the cost term is introduced into the weight update equation of the  SAMME.R algorithm, the updated data distribution is affected by the cost term $C_{ij}$. If the weight update parameter is not reintroduced, i.e., the cost term is considered for each cost-sensitive boosting algorithm, the boosting efficiency is not guaranteed. To justify the above scheme, we give the proof below. The core idea of our proof is that by minimizing the overall training error of the combined classifier, we generalize the weight update parameters for the proposed algorithm and derive an upper bound on the cumulative training error classification cost.

\newtheorem{theorem}{Theorem}
\begin{theorem}
The following holds for the upper bound of the training cumulative misclassification cost. $\mathbb{I}(\mathbb{\pi})$ returns 1 if predicate $\mathbb{\pi}=\text{true}$ or 0 otherwise.
\begin{align}
\sum_{i} C_{i} & \mathbb{I}\left(H\left(x_{i}\right) \neq y_{i}\right) \leqslant n K^{L} d \cdot \prod_{l=1}^{L} Z_{l} ,\nonumber\\
&where \quad d=\sum_{i} \frac{D^{(L+1)}\left(x_{i}\right)}{C_{i}^{L-1}}
\label{theorem}
\nonumber
\end{align}
$Z_l$ is a normalization factor chosen so that $D_{l+1}$ will be a distribution.
\end{theorem}

\begin{proof}
From Eq. (\ref{ensemble classification}), we kown that
\begin{equation}
H(v)=\arg \max _{k} \sum_{l=1}^{L} h_{k}^{(l)}(v)
\nonumber
\end{equation}

Now, combining Eq. (\ref{weak classifier}), we let 
\begin{align}
f_k(x_i)&=\sum_{l=1}^{L}h_{k}^{(l)}(v_i) \nonumber\\
&=\sum_{l=1}^{L}(K-1)\left(\log p_{k}^{(l)}\left(x_{i}\right)-\frac{1}{K} \sum_{k^{\prime}} \log p_{k^{\prime}}^{(l)}\left(x_{i}\right)\right)
\end{align}

In this method, the K-dimensional vector $\mathbf{f}\left(x_{i}\right)$ composed of $f_k(x_i)(where \ k=1, \ldots, K )$ can be represented as follows:
\begin{align}
\mathbf{f}\left(x_{i}\right) =  \left[\sum_{l = 1}^{L}(K-1)\left(\log p_{1}^{(l)}\left(x_{i}\right)-\frac{1}{K} \sum_{k^{\prime}} \log p_{k^{\prime}}^{(l)}\left(x_{i}\right)\right)\right. , \nonumber \\
\sum_{l = 1}^{L}(K-1)\left(\log p_{2}^{(l)}\left(x_{i}\right)-\frac{1}{K} \sum_{k^{\prime}} \log p_{k^{\prime}}^{(l)}\left(x_{i}\right)\right), \nonumber \\
\vdots    \qquad \qquad \qquad \qquad  \nonumber \\
\left.\sum_{l = 1}^{L}(K-1)\left(\log p_{K}^{(l)}\left(x_{i}\right)-\frac{1}{K} \sum_{k^{\prime}} \log p_{k^{\prime}}^{(l)}\left(x_{i}\right)\right)\right]^T
\label{fxi}
\end{align}

Similar to vanilla SAMME.R, we recode the label of each node with a $K$-dimensional vector $\mathbf{y}$, where all entries equal to $-\frac{1}{K-1}$ except a 1 in position $k$ if $c = k$, i.e. $\mathbf{y} =(y_1,...,y_K )$, and
\begin{equation}
y_{k} = \left\{\begin{array}{ll}
1, & \text { if } c = k \\
-\frac{1}{K-1}, & \text { if } c \neq k .
\end{array}\right.
\label{recodey}
\end{equation}

Therefore,  combining Eq. (\ref{fxi}) and (\ref{recodey}), we can get the inner product of $\mathbf{y}_{i}$ and $ \mathbf{f}\left(x_{i}\right)$ as:
\begin{align}
-&\mathbf{y}_{i} \mathbf{f}\left(x_{i}\right) \nonumber \\
=&\sum_{k \neq c} \sum_{l=1}^{L}\left(\log p_{k}^{(l)}\left(x_{i}\right)-\frac{1}{K} \sum_{k^{\prime}} \log p_{k^{\prime}}^{(l)}\left(x_{i}\right)\right) \nonumber \\
&-\sum_{l=1}^{L}(K-1)\left(\log p_{c}^{(l)}\left(x_{i}\right)-\frac{1}{K} \sum_{k^{\prime}} \log p_{k^{\prime}}^{(l)}\left(x_{i}\right)\right) \nonumber \\
=& \sum_{l = 1}^{L} \sum_{k \neq c}\left(\log p_{k}^{(l)}\left(x_{i}\right)-\log \left(\prod_{k^{\prime}} p_{k^{\prime}}^{(l)}\left(x_{i}\right)\right)^{\frac{1}{K}}\right) \nonumber\\
&-\sum_{l = 1}^{L}(K-1) \cdot\left(\log p_{c}^{(l)}\left(x_{i}\right)-\log \left(\prod_{k^{\prime}} p_{k^{\prime}}^{(l)}\left(x_{i}\right)\right)^{\frac{1}{K}}\right) \nonumber\\ 
=& \sum_{l = 1}^{L}\left(\sum_{k \neq c} \log p_{k}^{(l)}\left(x_{i}\right)-(K-1) \log \left(\prod_{k^{\prime}} p_{k^{\prime}}^{(l)}\left(x_{i}\right)\right)^{\frac{1}{K}}\right) \nonumber\\
&-\sum_{l = 1}^{L}(K-1) \cdot\left(\log p_{c}^{(l)}\left(x_{i}\right)-\log \left(\prod_{k^{\prime}} p_{k^{\prime}}\left(x_{i}\right)\right)^{\frac{1}{K}}\right) \nonumber \\ 
=& \sum_{l = 1}^{L}\left(\sum_{k \neq c} \log p_{k}^{(l)}\left(x_{i}\right)-\log \left(\prod_{k^{\prime}} p_{k^{\prime}}^{(l)}\left(x_{i}\right)\right)^{K-1}\right) \nonumber \\
=&\sum_{l=1}^{L} \log \frac{\prod_{k \neq c} p_{k}^{(l)}\left(x_{i}\right)}{\left(p_{c}^{(l)}\left(x_{i}\right)\right)^{K-1}}
\label{yf}
\end{align}

Taking the exponent for both sides of Eq. (\ref{yf}), we can obtain
\begin{align}
e^{-\mathbf{y}_{i} \mathbf{f}^{\prime}\left(x_{i}\right) }&=\prod_{l=1}^{L} e^{\log \frac{\prod_{k \neq c} p_{k}^{(l)}\left(x_{i}\right)}{\left(p_{c}^{(l)}\left(x_{i}\right)\right)^{K-1}}}  \nonumber\\
&=\prod_{l=1}^{L} \frac{\prod_{k \neq c} p_{k}^{(l)}\left(x_{i}\right)}{\left(p_{c}^{(l)}\left(x_{i}\right)\right)^{K-1}}
\label{expyf}
\end{align}

Unravelling the weight update rule of Eq. (\ref{node weight update}), using the iterative method and normalizing it with the normalization factor $Z_l$ , we can obtain
\allowdisplaybreaks[4]
\begin{align}
D&^{(L+1)}\left(x_{i}\right)  \nonumber\\
=& D^{(L)}\left(x_{i}\right) \cdot \frac{\left(-\frac{K-1}{K} \cdot C_i \cdot \mathbf{y}_{i} \cdot \log \mathbf{p}^{(L)}\left(x_{i}\right)\right)}{Z_l}  \nonumber\\
=& \frac{1}{n} \cdot\left(-\frac{K-1}{K}C_i\right)^{L} \Bigg( \prod_{l=1}^{L}\Bigg(\sum_{k \neq c}\left(-\frac{1}{K-1}\right) \log p_{k}^{(l)}\left(x_{i}\right) \Bigg.\Bigg. \nonumber\\
&\Bigg.\Bigg.+\log p_{c}^{(l)}\left(x_{i}\right)\Bigg) \Bigg) \cdot \frac{ 1}{\prod_{t=1}^{L}Z_l} \nonumber\\
=& \frac{1}{n} \cdot\left(-\frac{K-1}{K}C_i\right)^{L} \cdot \prod_{l=1}^{L} \Bigg( \left(-\frac{1}{K-1}\right) \cdot \log \prod_{k \neq c} p_{k}^{(l)}\left(x_{i}\right) \Bigg.\nonumber\\
&\Bigg.+\log p_{c}^{(l)}\left(x_{i}\right)\Bigg)\cdot \frac{1}{\prod_{l=1}^{L}Z_l}  \nonumber\\
=& \frac{1}{n} \cdot\left(\frac{1}{K}C_i \right)^{T} \cdot \Bigg(\prod_{l=1}^{L}\Bigg(\log _{k \neq c} \prod_{k} p_{k}^{(l)}\left(x_{i}\right)\Bigg. \Bigg.\nonumber\\
&\Bigg.\Bigg. -(K-1) \log p_{c}^{(l)}\left(x_{i}\right)\Bigg)\Bigg) \cdot \frac{1}{\prod_{l=1}^{L}Z_l}  \nonumber\\
=& \frac{1}{n}\left(\frac{1}{K}C_i \right)^{L} \cdot \frac{\prod_{l=1}^{L}\left(\log \frac{\prod_{k \neq c} p_{k}^{(l)}\left(x_{i}\right)}{\left(p_{c}^{(l)}\left(x_{i}\right)\right)^{K-1}}\right)}{\prod_{l=1}^{L}Z_l}
\end{align}

Rearranging the above equation, we can obtain
\begin{align}
\prod_{l=1}^{L}\left(\log \frac{\prod_{k \neq c} p_{k}^{(l)}\left(x_{i}\right)}{\left(p_{c}^{(l)}\left(x_{i}\right)\right)^{K-1}}\right)=n \cdot \left(\frac{K}{C_i}\right)^{L} \cdot D^{(L+1)}\left(x_{i}\right)\prod_{l=1}^{L}Z_l
\label{multilog}
\end{align}

Furthermore, when $H\left(x_{i}\right) \neq y_{i}$, $ y_iH(x_i)<0$. Therefore, $e^{-\mathbf{y}_{i} \mathbf{f}^{\prime}\left(x_{i}\right)}\ge 1$.When $H\left(x_{i}\right) = y_{i}$, $ y_iH(x_i)>0$. Therefore, $e^{-\mathbf{y}_{i} \mathbf{f}^{\prime}\left(x_{i}\right)}\ge 0$. In summary, the following inequality always holds:
\begin{equation}
\mathbb{I}\left(H\left(x_{i}\right) \neq y_{i}\right) \leqslant e^{-\mathbf{y}_{i} \mathbf{f}\left(x_{i}\right)}
\label{index inequal}
\end{equation}

Bringing Eq.(\ref{expyf}) into Eq. (\ref{index inequal}) , and multiplying both sides of the equation by $C_i$, we get the following inequality:
\begin{equation}
C_i\mathbb{I}\left(H\left(x_{i}\right) \neq y_{i}\right) \leqslant C_i\prod_{l=1}^{L} \frac{\prod_{k \neq c} p_{k}^{(l)}\left(x_{i}\right)}{\left(p_{c}^{(l)}\left(x_{i}\right)\right)^{K-1}}
\label{Cindex inequal}
\end{equation}

Combining the right-hand sides of Eq. (\ref{index inequal}) and Eq. (\ref{Cindex inequal}), bringing in Eq. (\ref{multilog}), and summing over all nodes, we get
\begin{align}
\sum_{i} C_{i} e^{-\mathbf{y}_{i} \cdot \mathbf{f}\left(x_{i}\right)} &=\sum_{i} C_{i} \prod_{l=1}^{L} \frac{\prod_{k \neq c} p_{k}^{(l)}\left(x_{i}\right)}{\left(p_{c}^{\left(l)\right.}\left(x_{i}\right)\right)^{K-1}} \nonumber \\
&=\sum_{i} n \cdot \frac{K^{L}}{C_{i}^{L-1}} \cdot D^{(L+1)}\left(x_{i}\right) \cdot \prod_{l=1}^{L} Z_{l} \nonumber \\
&=n \cdot K^{L} \cdot \prod_{l=1}^{L} Z_{l} \sum_{i} \frac{D^{(L+1)}\left(x_{i}\right)}{C_{i}^{L-1}} \nonumber \\
&=n K^{L} d \cdot \prod_{l=1}^{L} Z_{l}
\end{align}

So, we can get
\begin{equation}
\sum_{i} C_{i} \mathbb{I}\left(H\left(x_{i}\right) \neq y_{i}\right) \leqslant n K^{L} d \cdot \prod_{l=1}^{L} Z_{l} 
\label{theorem}
\nonumber
\end{equation}
where $ d=\sum_{i} \frac{D^{(L+1)}\left(x_{i}\right)}{C_{i}^{L-1}}$.

\end{proof}

\IncMargin{1em}
\begin{algorithm} [ht] \SetKwData{Left}{left}\SetKwData{This}{this}\SetKwData{Up}{up} \SetKwFunction{Union}{Union}\SetKwFunction{FindCompress}{FindCompress} \SetKwInOut{Input}{input}\SetKwInOut{Output}{output}

	\KwIn{An undirected graph with node features and labels: $\mathcal{G}=(\mathcal{V},\mathcal{X},\mathcal{A},\mathcal{E},\mathcal{Y})$ \;
	\quad \quad \quad The number of node classes $K$\;
	\quad \quad \quad Number of layers, epochs: $L,E$\;
	  }

	\KwOut{Classification result $C(v)$, $v \in \mathcal{V}_{train}$.}

	 \BlankLine 
	 \ \tcp*[h]{Initialization}\; 
	 \ $ w_v^{(1)}=1/N$, $ {\forall} v \in \mathcal{V}_{train}$; $\mathbf{X}^{(0)}=\mathbf{X}$\;
	 \ \tcp*[h]{ Train GAT-COBO}\;
     \For {$l = 1,2...L$}
     {
		 \For{$e=1,2,...E$}
		 { 
		     \ \tcp*[h]{Train weak classifier}\;
		 		 \  Forward propagation $\mathbf{h}_{v}^{\prime} \leftarrow $ Eq.(\ref{innerGAT embedding1}) or Eq.(\ref{innerGAT embedding2}), $ {\forall} v \in \mathcal{V}_{train}$ \; 
         \ Calculate GAT loss $\mathcal{L}_{GAT} \leftarrow $ Eq.(\ref{inter loss})  \;
         \ Calculate attention matrix $\mathbf{\Omega}^{(l)}\leftarrow$ Eq.(\ref{Omega}) \;
         \ \tcp*[h]{Feature update}\;
         \ Update the input features for the next weak classifier $\mathbf{X}^{(l)}\leftarrow$ Eq.(\ref{input feature})  \;
         \ Calculate the probability vector $p_k^{(l)}(v) \leftarrow$  Eq.(\ref{pkv}), $ {\forall} v \in \mathcal{V}_{train}$ \;
         \ Calculate feature update loss $\mathcal{L}_{MLP} \leftarrow $ Eq.(\ref{cross-entropy loss}) \;
		    \ Calculate the overall loss of the model $\mathcal{L}_{GAT-COBO} \leftarrow$ Eq.(\ref{model loss}) \;
		  } 
		\ \tcp*[h]{Cost sensitive learner}\;
    \ Prepare cost matrix $ \mathbf{C}\leftarrow$  Eq.(\ref{actual class}) or Eq.(\ref{log1p}) \;
	  \ Output of weak classifier $h_k^{(l)}(v) \leftarrow$  Eq.(\ref{weak classifier}), $\forall k \in \{1,\ldots K \},{\forall} v \in \mathcal{V}_{train}$ \;
	  \ Calculate classification cost $C_v^{(l)} \leftarrow$ Eq.(\ref{Cvpred}) \;
	  \ Update node weight $w_v^{(l+1)} \leftarrow $ Eq(\ref{node weight update})  \;
	  \ \text{Re-normalize $w_v^{(l+1)}$} \;
     }
     \ Ensemble classification result $ H(v) \leftarrow  $ Eq.(\ref{ensemble classification})\;
     
 	 	  \caption{GAT-COBO: Cost-Sensitive Graph Neural Network for Telecom Fraud Detection}
 	 	  \label{algorithm} 
 	 \end{algorithm}
 \DecMargin{1em}

\subsection{Proposed GAT-COBO}\label{Propose GAT-COBO}
Given an imbalanced fraud detection graph $\mathcal{G}$ and a set of training nodes $\mathcal{V}_{train}$, we train the weak classifier to minimize the cross-entropy loss:
\begin{equation}
\mathcal{L}_{GAT-COBO}=w_v\left(\mathcal{L}_{GAT}+\lambda_{1} \mathcal{L}_{MLP}\right)+\lambda_{2}\|\theta\|_{2}
\label{model loss}
\end{equation}
where $\lambda_{1}$ and $\lambda_{2}$ are the weight parameters, and $\|\theta\|_{2}$ is the $L_2$-norm of all model parameters. $w_{v}$ represents the sampling weight defined in Eq.(\ref{node weight update}).

The complete training algorithm is summarized in Algorithm \ref{algorithm}. First, we feed the node features and adjacency matrix to the GAT-based weak classifier to get the learned attention matrix. Then the attention matrix and node features are input into MLP for feature update. After softmax transformation, the node class probability $p_k^{(l)}(v)$ is obtained. We input it into the Boosting-based cost sensitive learner and calculate the classification result $h_v$. Based on this, the  misclassification cost at this layer can be obtained. The calculated cost is then used to calculate the updated sampling weights. We use this weight to constrain the classification loss of the next GAT-based weak classifier, so that the misclassified nodes from previous weak classi-
fier have a higher probability of being correctly classified by the next one. Finally, we sum all the layers to get the final cost-sensitive classification result $H(v)$.

\begin{table*}[htb] \centering
\caption{Dataset and graph statistics.}
\renewcommand\arraystretch{1.5} 
\begin{tabular}{|c|c|c|c|c|c|}
\hline
\textbf{Dataset name}             & \textbf{Nodes (fraud ratio)} & \textbf{Edges} & \textbf{Number of classes} & \textbf{Feature dimension} & \textbf{IR}             \\ \hline
\multirow{2}{*}{\textbf{Sichuan}} & \multirow{2}{*}{6106 (32.1\%)}                 & \multirow{2}{*}{838528}  & Benign: 4144                  & \multirow{2}{*}{55}        & \multirow{2}{*}{0.4735} \\ \cline{4-4}
                                  &                                                &                          & Fraud: 1962                   &                            &                         \\ \hline
\multirow{3}{*}{\textbf{BUPT}}    & \multirow{3}{*}{116,383 (7.3\%)}               & \multirow{3}{*}{350751}  & Benign: 99861                 & \multirow{3}{*}{39}        & \multirow{3}{*}{0.0809} \\ \cline{4-4}
                                  &                                                &                          & Fraud : 8448              &                            &                         \\ \cline{4-4}
                                  &                                                &                          & Courier : 8074                &                            &                         \\ \hline
\end{tabular}
\label{dataset}
\end{table*}

\section{Experiments}\label{experiment}
This section will present the performance of GAT-COBO on the imbalanced telecom fraud detection dataset and comparisons with the baseline methods. It mainly answers the following research questions.
\begin{itemize}
\item
\textbf{RQ1}: Does GAT-COBO outperform the state-of-the-art methods for graph-based anomaly detection?
\item
\textbf{RQ2}: How does GAT-COBO perform with respect to the graph imbalance problem?
\item
\textbf{RQ3}: What is the performance with respect to GNN over-smoothing problem?
\item
\textbf{RQ4}: What is the hyperparameter sensitivity and its impact on model design?
\item
\textbf{RQ5}: What is the computational complexity of the proposed model?
\end{itemize}

\subsection{Experimental setup}
\subsubsection{Dataset}
We use two real-world telecom fraud detection datasets to validate the proposed method. The first one\footnote{https://aistudio.baidu.com/aistudio/datasetdetail/40690.} is released in the 2020 Sichuan Big Data Competition. It was collected and anonymized by China Mobile Sichuan Company, which covers the CDR data of 6,106 users in 23 cities of Sichuan Province, with a time span of August 2019 to March 2020. The content includes call records (call object, duration, type, time, location), text message records (SMS object, text message type, communication time), Internet access records (traffic consumption, APP name), expense records (monthly consumption amount, attribution), etc. We performed raw data processing in the same way as \cite{hu2022btg}. All samples are divided into two categories, namely fraudsters and the benign. In this dataset, the imbalance rate $IR=1962/4144=0.4735$. 

The second dataset\footnote{https://github.com/khznxn/TF-Dataset.} was released in 2019 by Ming Liu et al. \cite{liu2019agrm} of Beijing University of Posts and Telecommunications. The dataset includes one week’s CDR data of users in a Chinese city. The author performed feature extraction on the original CDR data, and obtained a total of 39-dimensional features. According to the definition in Section 3, the imbalance rate of this dataset is $IR=8074/99861=0.0809$. Detailed descriptions are summarized in Table \ref{dataset}.


\subsubsection{Baselines}
To verify the learning ability of GAT-COBO on imbalanced graph data, we compare it with various GNN baselines in a semi-supervised learning setting. We choose GCN \cite{kipf2016semi}, GAT \cite{velivckovic2017graph}, GraphSAGE \cite{hamilton2017inductive} as general GNN models. We choose FdGars \cite{wang2019fdgars}, Player2Vec \cite{zhang2019key}, GraphConsis \cite{liu2020alleviating}, GEM \cite{liu2018heterogeneous} and CARE-GNN \cite{dou2020enhancing} as state-of-the-art GNN-based fraud detectors.

\begin{itemize}
\item
GCN: GNN that aggregates neighbor information by spectral graph convolution
\item
GAT: GNN that uses an attention mechanism to aggregate neighbor node information
\item
GraphSAGE: An inductive GNN with a fixed number of sampled neighbors
\item
FdGars: A GCN-based opinion fraud detection system
\item
Player2vec: A GNN for modeling heterogeneous relationships of homogeneous nodes using heterogeneous information networks and meta-Paths
\item
GraphConsis : A heterogeneous graph neural network for graph inconsistency
\item
GEM: A heterogeneous graph fraud detection GNN based on attention mechanism
\item
CARE-GNN: A heterogeneous GNN with multimodal aggregation based on reinforcement learning neighbor selection
\item
GAT-COBO$_{uni}$: our proposed method in which the cost matrix is calculated in \textbf{uniform} manner
\item
GAT-COBO$_{inv}$: our proposed method in which the cost matrix is computed in \textbf{inverse} manner
\item
GAT-COBO$_{log}$: our proposed method in which the cost matrix is calculated in \textbf{log1p} manner
\end{itemize}

\begin{table*}
\caption{Performance comparison on two real-world telecom fraud datasets.}
\renewcommand\arraystretch{1.5} 
\begin{tabular}{|c|c|cccc|cccc|}
\hline
\multirow{2}{*}{Method}                                                       & \textbf{Dataset}                                                      & \multicolumn{4}{c|}{\textbf{Sichuan}}                                                                                                & \multicolumn{4}{c|}{\textbf{BUPT}}                                                                                                   \\ \cline{2-10} 
                                                                              & \textbf{Metric}                                                       & \multicolumn{1}{c|}{Macro AUC}       & \multicolumn{1}{c|}{Macro recall}    & \multicolumn{1}{c|}{Macro F1}        & G-mean        & \multicolumn{1}{c|}{Macro AUC}       & \multicolumn{1}{c|}{Macro recall}    & \multicolumn{1}{c|}{Macro F1}        & G-mean       \\ \hline
\multirow{3}{*}{\begin{tabular}[c]{@{}c@{}}General\\ GNNs\end{tabular}}       & \textbf{GCN}                                                          & \multicolumn{1}{c|}{0.9263}          & \multicolumn{1}{c|}{0.8597}          & \multicolumn{1}{c|}{0.8755}          & 0.8530          & \multicolumn{1}{c|}{0.8932}          & \multicolumn{1}{c|}{0.5706}          & \multicolumn{1}{c|}{0.6265}          & 0.4380          \\ \cline{2-10} 
                                                                              & \textbf{GAT}                                                          & \multicolumn{1}{c|}{0.9243}          & \multicolumn{1}{c|}{0.8585}          & \multicolumn{1}{c|}{0.8725}          & 0.8529          & \multicolumn{1}{c|}{0.9102}          & \multicolumn{1}{c|}{0.6152}          & \multicolumn{1}{c|}{0.6803}          & 0.5267          \\ \cline{2-10} 
                                                                              & \textbf{Graphsage}                                                    & \multicolumn{1}{c|}{0.9159}          & \multicolumn{1}{c|}{0.8564}          & \multicolumn{1}{c|}{0.8631}          & 0.8447          & \multicolumn{1}{c|}{0.8928}          & \multicolumn{1}{c|}{0.6715}          & \multicolumn{1}{c|}{0.6918}          & 0.5823          \\ \hline
\multirow{5}{*}{\begin{tabular}[c]{@{}c@{}}GNN-\\based\\  methods\end{tabular}} & \textbf{FdGars}                                                       & \multicolumn{1}{c|}{0.7887}          & \multicolumn{1}{c|}{0.7082}          & \multicolumn{1}{c|}{0.6499}          & 0.6914          & \multicolumn{1}{c|}{0.6462}          & \multicolumn{1}{c|}{0.4357}          & \multicolumn{1}{c|}{0.4027}          & 0.3855          \\ \cline{2-10} 
                                                                              & \textbf{Player2vec}                                                   & \multicolumn{1}{c|}{0.7467}          & \multicolumn{1}{c|}{0.5618}          & \multicolumn{1}{c|}{0.4097}          & 0.4181          & \multicolumn{1}{c|}{0.5227}          & \multicolumn{1}{c|}{0.3206}          & \multicolumn{1}{c|}{0.3239}          & 0.2502          \\ \cline{2-10} 
                                                                              & \textbf{GraphConsis}                                                  & \multicolumn{1}{c|}{0.7985}          & \multicolumn{1}{c|}{0.7288}          & \multicolumn{1}{c|}{0.7331}          & 0.7187          & \multicolumn{1}{c|}{0.6211}          & \multicolumn{1}{c|}{0.3311}          & \multicolumn{1}{c|}{0.3074}          & 0.1602          \\ \cline{2-10} 
                                                                              & \textbf{GEM}                                                          & \multicolumn{1}{c|}{0.8619}          & \multicolumn{1}{c|}{0.8209}          & \multicolumn{1}{c|}{0.8294}          & 0.8153          & \multicolumn{1}{c|}{0.6788}          & \multicolumn{1}{c|}{0.3344}          & \multicolumn{1}{c|}{0.3100}          & 0.0117          \\ \cline{2-10} 
                                                                              & \textbf{CARE-GNN}                                                      & \multicolumn{1}{c|}{0.9384}          & \multicolumn{1}{c|}{0.8717}          & \multicolumn{1}{c|}{0.8659}          & 0.8711          & \multicolumn{1}{c|}{0.9065}          & \multicolumn{1}{c|}{0.7642}          & \multicolumn{1}{c|}{0.5345}          & 0.7538          \\ \hline
\multirow{3}{*}{Ours}                                                         & \textbf{\begin{tabular}[c]{@{}c@{}}GAT-COBO$_{uni}$ \end{tabular}} & \multicolumn{1}{c|}{0.9292}          & \multicolumn{1}{c|}{0.8880}          & \multicolumn{1}{c|}{0.9014}          & 0.8844          & \multicolumn{1}{c|}{0.9041}          & \multicolumn{1}{c|}{0.7015}          & \multicolumn{1}{c|}{0.7428}          & 0.6572          \\ \cline{2-10} 
                                                                              & \textbf{\begin{tabular}[c]{@{}c@{}}GAT-COBO$_{inv}$\end{tabular}} & \multicolumn{1}{c|}{0.9385}          & \multicolumn{1}{c|}{\textbf{0.8894}} & \multicolumn{1}{c|}{\textbf{0.9022}} & \textbf{0.8860} & \multicolumn{1}{c|}{0.8919}          & \multicolumn{1}{c|}{0.7590}          & \multicolumn{1}{c|}{\textbf{0.7564}} & 0.7421          \\ \cline{2-10} 
                                                                              & \textbf{\begin{tabular}[c]{@{}c@{}}GAT-COBO$_{log} $\end{tabular}}   & \multicolumn{1}{c|}{\textbf{0.9391}} & \multicolumn{1}{c|}{0.8867}          & \multicolumn{1}{c|}{0.9003}          & 0.8829          & \multicolumn{1}{c|}{\textbf{0.9109}} & \multicolumn{1}{c|}{\textbf{0.7823}} & \multicolumn{1}{c|}{0.7557}          & \textbf{0.7658} \\ \hline
\end{tabular}
\label{overalltable}
\end{table*}

\subsubsection{Experiment implementation}
In the experiments, we randomly select training samples and keep the ratio of positive and negative samples in the training set the same as the whole dataset. For our proposed GAT-COBO method, we use Adam optimizer for parameter optimization, and the specific configuration is as follows. For the Sichuan dataset, we set hid embedding size (128), learning rate (0.002), dropout (0), adj\_dropout (0.4), model layers (2), attention loss weight (0.5), attention weight (0.1), feature weight (0.1). For the BUPT dataset, we set hid embedding size (64), learning rate (0.01), dropout (0.2), the adj\_dropout (0.1), model layers (2), the attention loss weight (0.01), the attention weight (0.5), feature weight (0.6).
For GCN, GAT, CARE-GNN, we use the open-source implementation of Deep Graph Library(DGL) \footnote{https://github.com/dmlc/dgl}. For GraphSAGE, FdGars, Player2Vec, GraphConsis, and GEM , we use the open-source implementation of DGFraud-TF2 \footnote{https://github.com/safe-graph/DGFraud-TF2}. We implement the proposed method by Pytorch; all models are running on python3.7.10, 1 GeForce RTX 3090 GPU, 64GB RAM, 16 cores Intel(R) Xeon(R) Gold 5218 CPU @2.30GHz Linux Server.

\subsubsection{Evaluation metrics}
For the imbalanced problems, the evaluation metric is very critical. Because it needs to reasonably assess the classification results of all classes, especially the minority, and reflect them in the scores \cite{liang2021credit}. To be unbiased, we adopt four widely used metrics to measure the performance of all comparison methods: Macro AUC, Marco F1, Macro recall, and G-mean.

\noindent\textbf{Recall} is very important for the imbalance problem, which can accurately measures the proportion of minority class being detected. It is defined as follows:
\begin{equation}
recall=\frac{TP}{TP+TN}
\label{recall}
\nonumber
\end{equation}
where $TP$ and $TN$ represent the number of true positive and true negative samples in the confusion matrix, respectively. Macro recall is the arithmetic mean of multiple classes, which treats all classes equally, regardless of the importance of different classes.

\noindent\textbf{F1} is another comprehensive matric for evaluating imbalanced problem, which is defined as follows:
\begin{equation}
F1=\frac{2}{\frac{1}{\text { precision }}+\frac{1}{\text { recall }}}=2 \frac{\text { precision } \times \text { recall }}{\text { precison }+\text { recall }}
\label{F1}
\nonumber
\end{equation}
and macro-f1 is to calculate the arithmetic mean of F1 scores for each class.

\noindent\textbf{AUC} is the area under the ROC curve, defined as:
\begin{equation}
\mathrm{AUC}=\frac{\sum_{u \in \mathcal{U}^{+}} \operatorname{rank}_{u}-\frac{\left|\mathcal{U}^{+}\right| \times\left(\left|\mathcal{U}^{+}\right|+1\right)}{2}}{\left|\mathcal{U}^{+}\right| \times\left|\mathcal{U}^{-}\right|}
\label{AUC}
\nonumber
\end{equation}

Here, $\mathcal{U}^{+}$ and $\mathcal{U}^{-}$ indicate the minority and majority class set in the testing set, respectively. And $rank_u$ indicates the rank of node $u$ via the score of prediction.

\noindent\textbf{G-Mean} calculates the geometric mean of True Positive Rate (TPR) and True Negative Rate (TNR).
\begin{equation}
\text { G-Mean }=\sqrt{\mathrm{TPR} \cdot \mathrm{TNR}}=\sqrt{\frac{\mathrm{TP}}{\mathrm{TP}+\mathrm{FN}} \cdot \frac{\mathrm{TN}}{\mathrm{TN}+\mathrm{FP}}}
\label{GMean}
\nonumber
\end{equation}

For the above metrics, the higher the score, the better the performance of the model on imbalanced problems.

\subsection{Performance comparison(RQ1)}
We evaluate the performance of all the comparison methods on the above mentioned telecom fraud detection datasets. And the best testing results of GAT-COBO and baseline methods for Macro AUC, Marco F1, Macro recall and G-mean metrics are reported in Table \ref{overalltable}. The compared methods can be further divided into three groups, namely general-purpose GNNs, GNN-based fraud detectors, our methods. All comparison methods use the same training set, validation set, and test set division, and the ratios are 20\%, 20\%, and 60\%, respectively. We observe that GAT-COBO outperforms other baselines under all metrics on both datasets. Moreover, the performance of the inverse and $log1p$ variants of the proposed method is significantly better than that of the $uniform$, which indicates that considering the imbalanced class cost is crucial to the performance of GAT-COBO on imbalanced datasets. In addition, we observe the following experimental results.

First of all, GCN, GAT, and GraphSAGE are three classic general GNNs, and have more moderate performance on both datasets. However, for such imbalanced fraud detection datasets, general GNNs cannot take into account the minority class. Because they treat all classes of nodes equally, which makes their performance lower than our proposed GAT-COBO. This difference in performance shows that GNN can perform better on imbalanced problems by designing sensitive to the cost of different classes.

In contrast, the performance of GNN-based fraud detectors varies greatly. CARE-GNN outperforms most similar methods and outperforms general GNNs on the two original datasets. From the G-Mean score, it can be observed that CARE-GNN is only lower than our method, which shows that it has strong adaptability to imbalanced datasets. This is mainly due to the selection of similar neighbors by the reinforcement learning module in the model, as well as mini-batch training and undersampling techniques. While other GNN-based fraud detectors perform poorly. The reasons are mainly two-fold, one is that these models do not take into account the class imbalance in the data. Another aspect is that they are designed for multi-relationship graphs, whereas both datasets in this paper are single-relational graphs.

Furthermore, by comparing the performance of all methods on the two fraud detection datasets, we can find that most methods perform much better on Sichuan than BUPT. An important reason is that the IR of Sichuan is larger than that of BUPT ( 0.4735$>$0.0809 ), which indicates that the class imbalance of BUPT is more serious than Sichuan. Moreover, the more unbalanced the dataset, the greater the performance gap of the same model, which shows the significant impact of data class imbalance on model performance. However, the performance of our method on these two datasets is not very different and both are at a high level. This further illustrates the important role played by this cost-sensitive design in our proposed method.

\begin{figure*}[ht]
\centering

\subfigure[Imbalance rate]{
\label{G-mean}
\includegraphics[width=5.5cm]{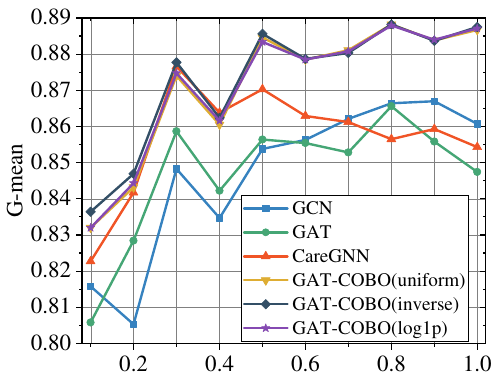}
}
\hspace{1mm}
\subfigure[Imbalance rate]{
\label{AUC}
\includegraphics[width=5.5cm]{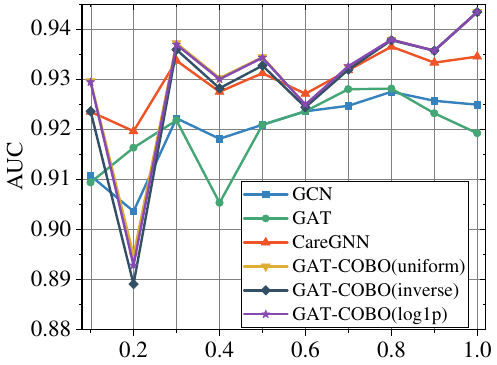}
}
\hspace{1mm}
\subfigure[Imbalance rate]{
\label{recall}
\includegraphics[width=5.5cm]{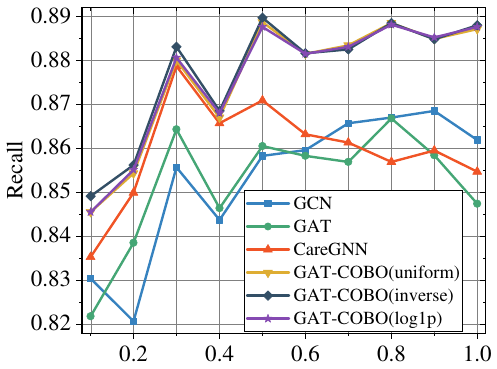}
}
\hspace{1mm}
\subfigure[Imbalance rate]{
\label{G-mean}
\includegraphics[width=5.5cm]{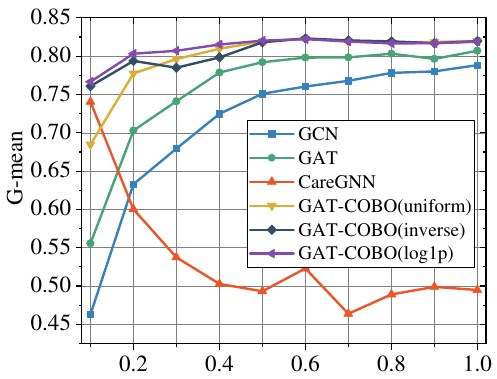}
}
\hspace{1mm}
\subfigure[Imbalance rate]{
\label{AUC}
\includegraphics[width=5.5cm]{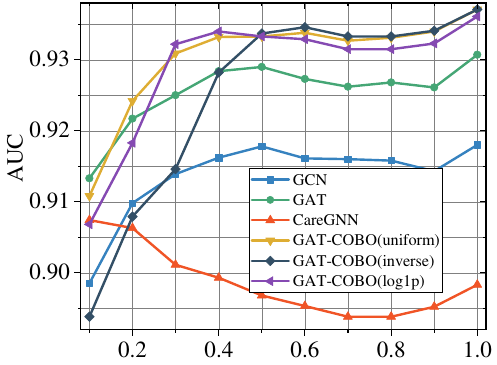}
}
\hspace{1mm}
\subfigure[Imbalance rate]{
\label{recall}
\includegraphics[width=5.5cm]{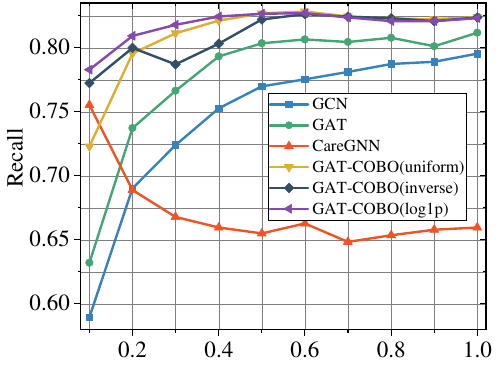}
}

\caption{Performance comparison of baseline methods with different IR. The top row is on the Sichuan dataset, and the bottom is on BUPT.}
\label {IR}
\end{figure*}

\subsection{Influence of imbalance ratio(RQ2)}
In this subsection, we test the performance of comparison algorithms under different imbalance rates to evaluate their robustness. In the experiment, we randomly sample different class of samples in the two telecom fraud detection datasets according to the IR (ranging from 0.1 to 1) to form a new dataset. Then we select three well-performing baselines in Table \ref{overalltable}, namely GCN, GAT, CARE-GNN and three variants of the GAT-COBO to test under the new dataset. Their scores on G-mean, AUC, macro Recall are recorded, and the result is shown in Figure \ref{IR}.
From the figure, we draw the following observations:

Our proposed method achieves the best results under almost all IRs, regardless of which metric. In particular, the $lop1p$ variant of GAT-COBO has an excellent performance in most situations. This fully demonstrates the effectiveness of the cost-sensitive boosting method in dealing with the problem of graph imbalance. In addition, we also notice that other methods decay faster as IR decreases, while our proposed method decays relatively slowly, which further illustrates the role of cost sensitivity when the classes are extremely unbalanced.

CARE-GNN is an effective GNN-based fraud detection model against fraudster camouflage, achieving SOTA on social datasets. However, we observe that its performance on two different telecom fraud datasets is quite different. In particular, on the BUPT dataset, the performance of CARE-GNN gradually decreases when $IR$ converges to 1. This may be due to the fact that the reinforcement learning-based sampling mechanism does not sample effective neighbors when the node neighbors gradually become more numerous. This suggests the limitation of the sampling method in solving the graph imbalance problem. It also gives us a lesson that GNN model needs to be specially designed when solving the graph imbalance problem. As classic GNN, GCN and GAT have stable performance and fast calculation speed, and can achieve good results in most scenarios. But they also lack the ability to deal with imbalanced data.

\subsection{Avoidance of over-smoothing effects(RQ3)}
\begin{figure*}[ht]
\centering

\subfigure[Layers]{
\label{G-mean}
\includegraphics[width=5.5cm]{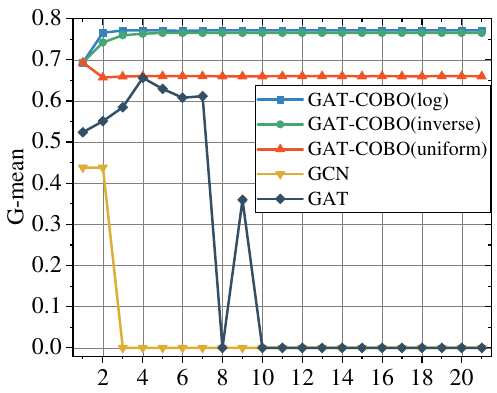}
}
\hspace{1mm}
\subfigure[Layers]{
\label{AUC}
\includegraphics[width=5.5cm]{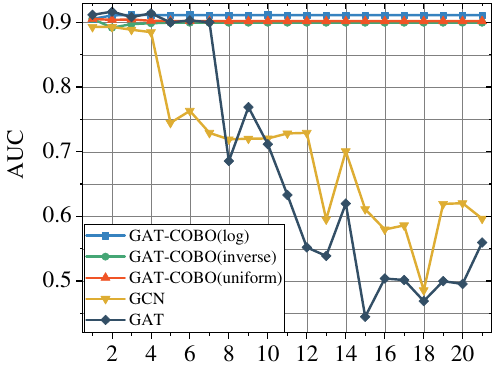}
}
\hspace{1mm}
\subfigure[Layers]{
\label{recall}
\includegraphics[width=5.5cm]{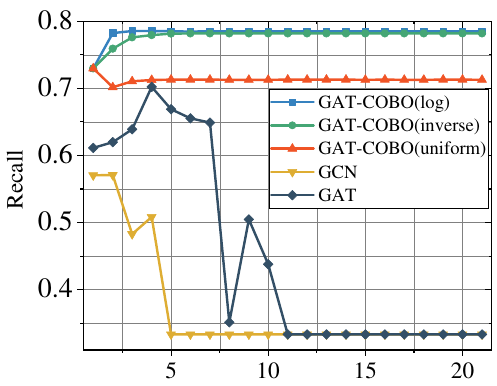}
}
\caption{Performance comparison of baseline methods with different GNN layers on BUPT dataset.}
\label {layers}
\end{figure*}

\begin{figure*}[ht]
\centering

\subfigure[Train size]{
\label{trainsize}
\includegraphics[width=5.5cm]{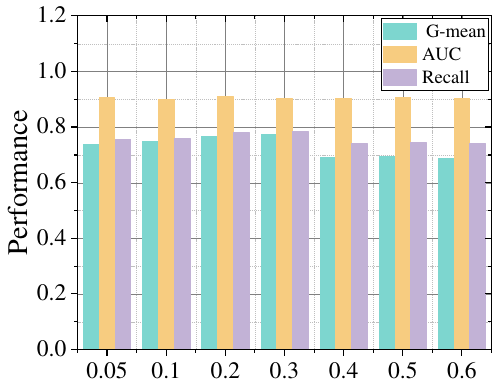}
}
\hspace{1mm}
\subfigure[Embedding size]{
\label{hidunit}
\includegraphics[width=5.5cm]{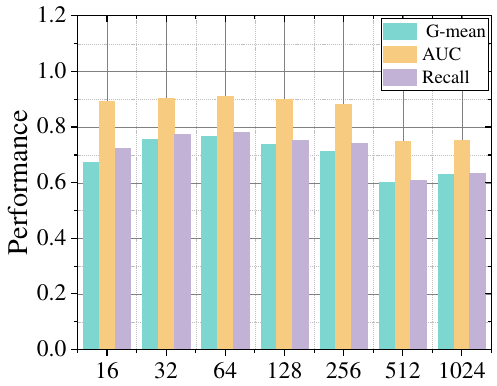}
}
\hspace{1mm}
\subfigure[Learning rate]{
\label{lr}
\includegraphics[width=5.5cm]{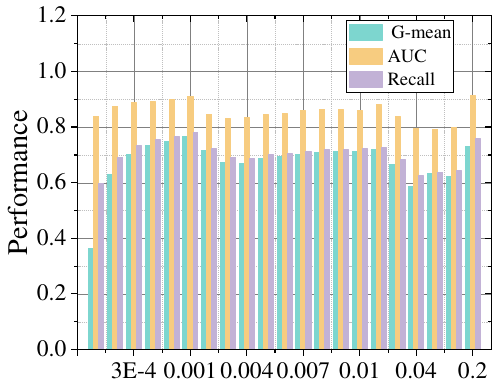}
}
\caption{ Hyperparameter sensitivity of the paoposed GAT-COBO.}
\label {HP}
\end{figure*}

It is well known that with the stacking of more graph operation layers, classic GNN models such as GCN and GAT will suffer from over-smoothing effects \cite{li2018deeper}. Different from the residual connection solution, in our method, we circumvent the over-smoothing effect through an ensemble learning method while improving the model performance. In order to verify the effect of our proposed method on the GNN over-smoothing problem, we compared the performance of the proposed method with GCN and GAT on G-mean, AUC, recall. During the experiment, we set the number of layers of GNN to vary from 1 to 21, and the results obtained are shown in Figure \ref{layers}.

From the figure, we can see that the performance of GCN and GAT is severely weakened by the over-smoothing effect. As the number of network layers increases, the performance of the models on all three metrics drops significantly. The over-smoothing effect of GCN is more severe than that of GAT. This is because the convolution operation treats all neighbors equally, while GAT partially mitigates the over-smoothing effect by weighting the neighbors differently. In contrast, the three different variants of GAT-COBO remain stable on all three metrics, and the model performance is largely unaffected by the number of network layers. It shows that our proposed method can effectively avoid the over-smoothing effect of GNN. In particular, the performance of the $inverse$ and $log1p$ variants increases slightly with the number of network layers. This is because we use ensemble learning that integrates multiple different weak learners, which allows the model to both maintain the initial information and learn the depth information.

\subsection{Hyperparameter Sensitivity(RQ4)}

In the hyperparameter sensitivity experiment, we tested the proposed model under the $log1p$ variant on the BUPT dataset. The three important hyperparameters are train size, embedding size, and learning rate. For each hyperparameter we recorded the model's score on G-mean, Macro AUC, Macro recall. The experimental results are shown in Figure \ref{HP}.

From Figure \ref{trainsize}, we observe that the model basically maintains stable performance as the training rate changes. Among them, the best effect is when the training rate is around 0.3. Therefore, we conclude that GAT-COBO is robust to training rate . Figure \ref{hidunit} shows the effect of different embedding sizes on the model. The model has comparable performance when the embedding sizes are 32, 64 and 128 . Figure \ref{lr} illustrates the effect of different learning rates on the model. It can be observed that when the learning rate is around 0.001 and 0.01, the model shows two small peaks in performance. Therefore, choosing an appropriate learning rate has a certain impact on model optimization.

\subsection{Computational Complexity(RQ5)}
Computational complexity is important for neural network models. This is because too high computational overhead can hinder model deployment in real-world scenarios, especially those that are more sensitive to timeliness. 

\begin{figure}[h]
\centering
  \includegraphics[width=8cm]{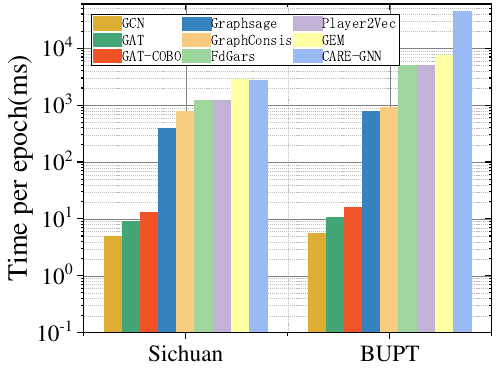}
  \caption{Per-epoch training time of GAT-COBO and baselines under 5 runs on two telecom fraud detection datasets.}
  \label{time}
\end{figure}

Figure \ref{time} shows the training time per epoch for GAT-COBO and the comparison baselines after five runs on two telecom fraud detection datasets (note that the vertical axis is in logarithmic coordinates). It can be observed that the computational complexity of GCN, GAT, and GAT-COBO is in the same order of magnitude, around 10 ms, while that of the remaining methods is around 1000 ms. The computational complexity of GAT-COBO is only slightly higher than that of GCN and GAT, and much lower than that of the other comparison baselines, which indicates that GAT-COBO requires less computational overhead. This does not seem to be consistent with our intuition, because GAT-COBO with the Boosting algorithm does not have much computational overhead. The reason for this phenomenon is that the computational overhead of the Boosting algorithm depends mainly on the base classifier. The base classifier of GAT-COBO uses a simplified GAT with low computational overhead, which contributes significantly to the overall computational complexity reduction of the model. The rest of the baselines with higher computational overhead mostly involve node neighbor sampling operations, such as Graphsage's neighbor sampling and CARE-GNN's reinforcement learning-based neighbor sampling. While these operations can reduce unnecessary information aggregation, they can significantly increase the model computational overhead. In addition, we also observe that the computational overhead of almost all models on BUPT is larger than that on Sichuan. This is because the number of nodes in BUPT is about 20 times larger than that in Sichuan, thus leading to a larger computational overhead of the models on BUPT.

\section{Conclusion}\label{conclusion}
 Graph imbalance problem can significantly affect the performance of telecom fraud detectors, but it's rarely noticed by previous work. In this paper, we propose a novel cost-sensitive graph neural network based on attention mechanism and ensemble learning to solve this problem. Concretely, we first learn node embeddings using graph attention network as base classifiers. The learned embeddings are then fed into the corresponding cost-sensitive learners for further training, and new node weights are calculated. This weight is then fed into the next GNN weak classifier as a constraint on the loss. Integrating the embeddings learned from such multiple base classifiers yields the final cost-sensitive classification result. We conduct extensive experiments on two real-world telecom fraud detection datasets to evaluate the proposed method. Experimental results demonstrate that the proposed GAT-COBO model outperforms state-of-the-art baseline methods and can effectively handle graph data with imbalanced class distributions. In addition, experiments also show that the proposed model can overcome the over-smoothing problem that is widespread in GNNs.
 Given that the graph imbalance problem is widely present in real-world tasks, GAT-COBO can be applied not only in telecom fraud detection, but also in scenarios such as social network bot user detection, financial fraud detection, and malicious machine detection, etc.


\section*{Acknowledgements}
We would like to thank the anonymous reviewers for their valuable comments.

\ifCLASSOPTIONcaptionsoff
  \newpage
\fi



%
%
%

\bibliographystyle{IEEEtran}
\bibliography{IEEEabrv,references}

%

\begin{IEEEbiography}[{\includegraphics[width=1in,height=1.25in,clip,keepaspectratio]{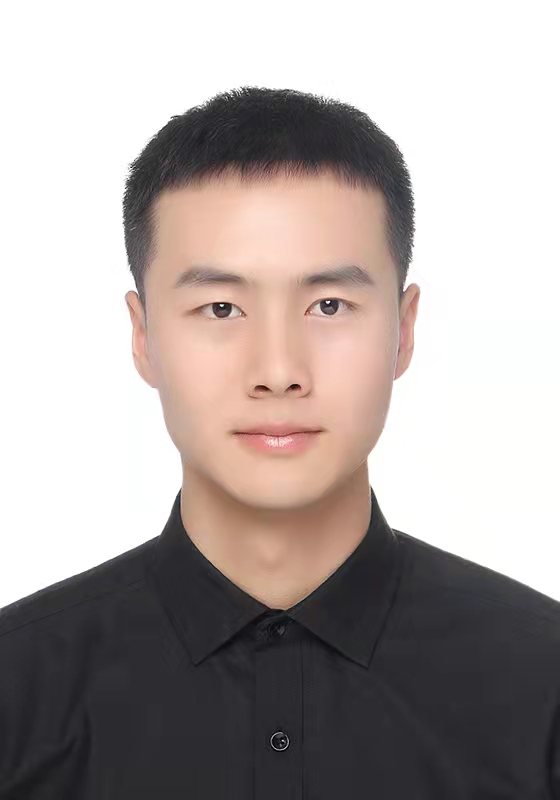}}]{Xinxin Hu}
 is currently a Doctor candidate at National Digital Switching System Engineering and Technological Research Center(NDSC) at Zhengzhou, China. He received the B.E. degree in Huazhong University of Science and Technology in 2017. His research interests include big data analysis, complex network, machine learning, mobile communication network security. 
\end{IEEEbiography}

\begin{IEEEbiography}[{\includegraphics[width=1in,height=1.25in,clip,keepaspectratio]{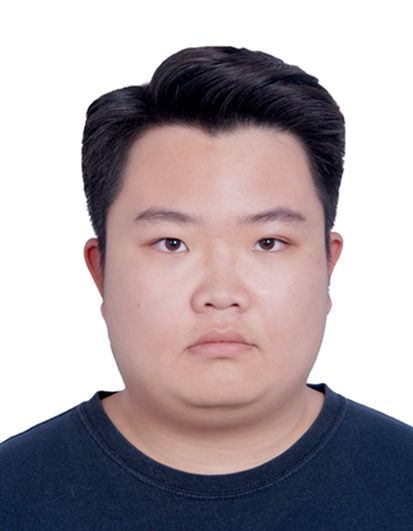}}]{Haotian Chen}
is currently a MEng student at The Edward S. Rogers Sr. Department of Electrical \& Computer Engineering, University of Toronto at Toronto, Canada. He received his Bachelor of computer science degree from University of Calgary. His research interests include machine learning, big data analysis, and network security.
\end{IEEEbiography}

\begin{IEEEbiography}[{\includegraphics[width=1in,height=1.25in,clip,keepaspectratio]{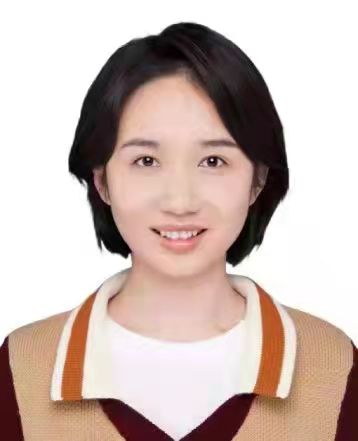}}]{Junjie Zhang}
is currently studying for a Ph.D. in National Digital Switching System Engineering Technology Research Center (NDSC). She received a B.E. degree from Xi'an Jiaotong University in 2018. Her research interests include complex networks, machine learning, and network robustness.
\end{IEEEbiography}


\begin{IEEEbiography}[{\includegraphics[width=1in,height=1.25in,clip,keepaspectratio]{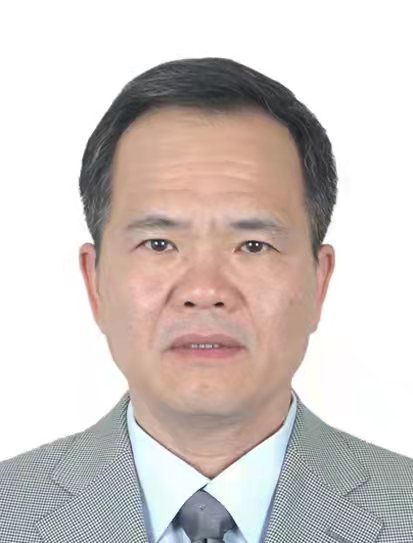}}]{Hongchang Chen}
is currently professor at National Digital Switching System Engineering and Technological Research Center(NDSC). His research interests include future network communication and future network architecture.
\end{IEEEbiography}

\begin{IEEEbiography}[{\includegraphics[width=1in,height=1.25in,clip,keepaspectratio]{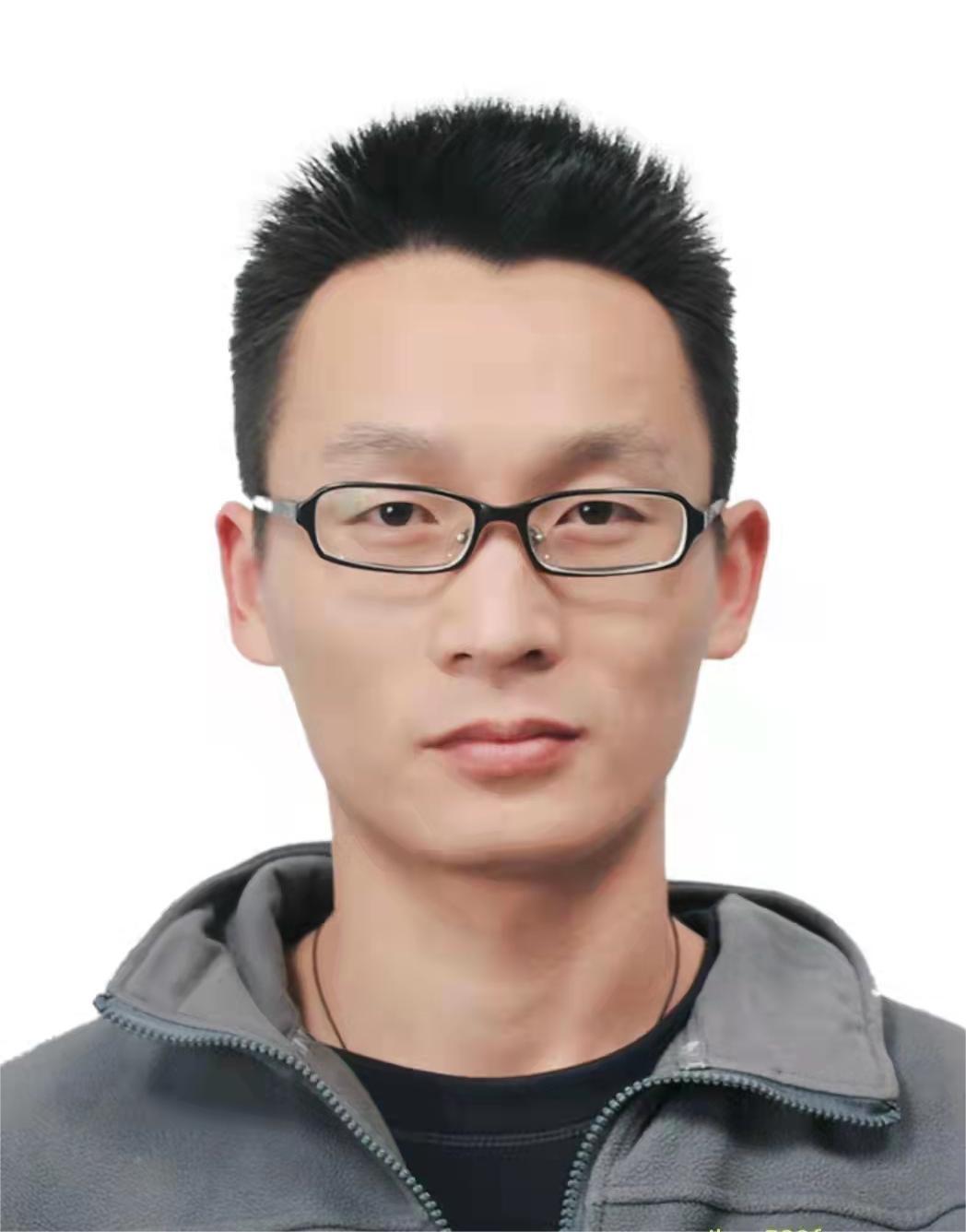}}]{Shuxin Liu}
is currently an Assistant Research Fellow with NDSC and the Director of the Laboratory of Network Architecture and Signaling Protocol Analysis. His research interests include network evolution, link prediction, network behavior analysis, and communication network security. 
\end{IEEEbiography}

\begin{IEEEbiography}[{\includegraphics[width=1in,height=1.25in,clip,keepaspectratio]{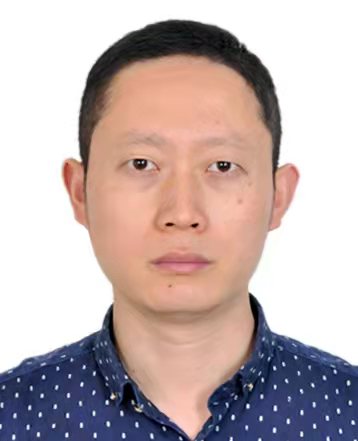}}]{Xing Li}
is  currently an assistant researcher of National Digital Switching System Engineering and Technological Research Center(NDSC), China. He received the Ph.D. degree from Information Engineering University in 2020. His major research interests include network science and cyberspace security.
\end{IEEEbiography}

\begin{IEEEbiography}[{\includegraphics[width=1in,height=1.25in,clip,keepaspectratio]{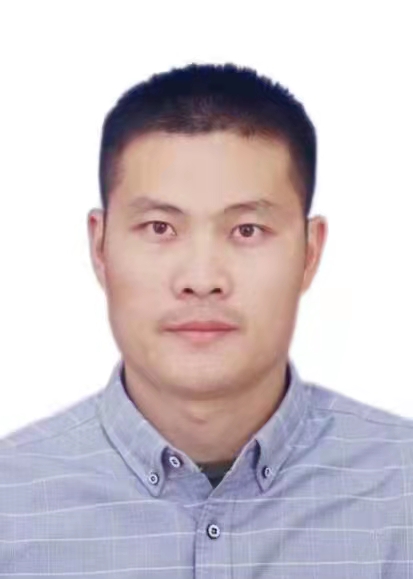}}]{Yahui Wang}
is currently a Doctor candidate at National Digital Switching System Engineering and Technological Research Center(NDSC) at Zhengzhou, China. He received the B.E. degree in Xi'an Jiaotong University in 2009. His research interests include data mining, complex network, machine learning.
\end{IEEEbiography}

\begin{IEEEbiography}[{\includegraphics[width=1in,height=1.25in,clip,keepaspectratio]{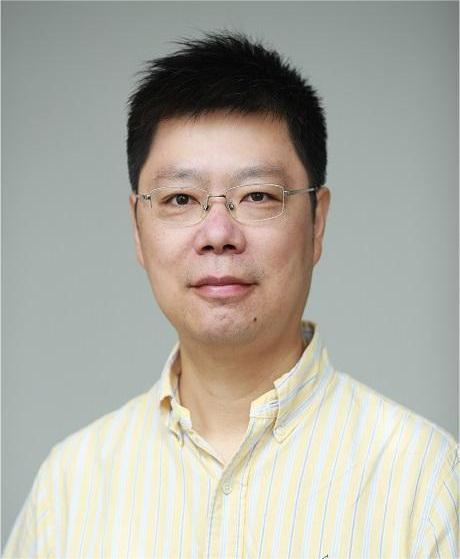}}]{Xiangyang Xue}
is currently a professor of computer science with Fudan University, Shanghai, China. He received the BS, MS, and PhD degrees in communication engineering from Xidian University, Xi'an, China, in 1989, 1992, and 1995, respectively. His research interests include multimedia information processing, big data analysis and machine learning. 
\end{IEEEbiography}




\end{document}